\documentclass{article}

\usepackage[final,nonatbib]{neurips_2020}

\usepackage[utf8]{inputenc} %
\usepackage[T1]{fontenc}    %
\usepackage{hyperref}       %
\usepackage{url}            %
\usepackage{booktabs}       %
\usepackage{amsfonts}       %
\usepackage{nicefrac}       %
\usepackage{microtype}      %

\usepackage{graphicx}
\usepackage{amsmath}
\usepackage{amssymb}
\usepackage{dblfloatfix}
\usepackage{transparent}

\usepackage{booktabs}
\usepackage{tabulary}
\usepackage[dvipsnames]{xcolor}

\usepackage{soul}
\usepackage{xspace}
\usepackage{adjustbox}
\usepackage{array}

\usepackage{algorithm}
\usepackage{listings}
\usepackage{algorithmic}
\usepackage{balance}

\def\x{{$\times$}}
\def\X{{$\times$}}

\usepackage[margin=4pt,font=small,labelfont=bf,labelsep=endash]{caption}

\usepackage{etoolbox}
\makeatletter
\AfterEndEnvironment{algorithm}{\let\@algcomment\relax}
\AtEndEnvironment{algorithm}{\kern2pt\hrule\relax\vskip3pt\@algcomment}
\let\@algcomment\relax
\newcommand\algcomment[1]{\def\@algcomment{\footnotesize#1}}
\renewcommand\fs@ruled{\def\@fs@cfont{\bfseries}\let\@fs@capt\floatc@ruled
  \def\@fs@pre{\hrule height.8pt depth0pt \kern2pt}%
  \def\@fs@post{}%
  \def\@fs@mid{\kern2pt\hrule\kern2pt}%
  \let\@fs@iftopcapt\iftrue}
\makeatother

\definecolor{codegreen}{rgb}{0,0.5,0}
\definecolor{codeblue}{rgb}{0.25,0.5,0.5}
\definecolor{codegray}{rgb}{0.6,0.6,0.6}

\usepackage{xspace}
\usepackage{bbm}
\usepackage{float} 
\usepackage{makecell}
\usepackage{subfigure}

\def\method{{\Ours}\xspace}
\def\OurMethod{{\Ours}\xspace}
\newcommand{\myparagraph}[1]{{\bf #1}}

\newcommand{\etal}{\textit{et al}.}
\newcommand{\ie}{\textit{i}.\textit{e}.}
\newcommand{\eg}{\textit{e}.\textit{g}.}
\usepackage{amssymb}%
\usepackage{pifont}%
\newcommand{\cmark}{\ding{51}}%
\newcommand{\xmark}{\ding{55}}%

\def\Ours{{SOLOv2}\xspace}

\title{
SOLOv2: Dynamic and Fast Instance Segmentation
}

\author{
        Xinlong Wang$ ^1 $
        ~~~~~~~~
        Rufeng Zhang$ ^{2} $ 
        ~~~~~~~~
        Tao Kong$ ^{3}$
        ~~~~~~~~
        Lei Li$ ^3  $
        ~~~~~~~~
        Chunhua Shen$ ^{1}$
        \\[0.152cm]
        $ ^1 $ The University of Adelaide, Australia 
        ~~~~
        $ ^2 $ Tongji University, China
        ~~~~
        $ ^3 $ ByteDance AI Lab  
}

\begin{document}

\maketitle

\begin{abstract}

In this work, we %
design
a simple, direct, and fast framework for instance segmentation with strong performance.
To this end, we propose a novel and effective approach, 
termed \method, following the principle of 
the SOLO method of Wang \textit{et al.}\ ``SOLO: segmenting  objects  by  locations'' \cite{solo}.
First, our new framework is empowered by an efficient and holistic instance mask representation scheme, which dynamically segments each instance in the image, without resorting to bounding box detection. 
Specifically, the object mask generation is decoupled into a mask kernel prediction and mask feature 
learning, 
which are responsible for generating %
convolution kernels and the %
feature maps to be convolved with, respectively. 
Second, \method significantly reduces inference overhead with our novel matrix non-maximum suppression (NMS) technique.
Our Matrix NMS performs NMS with parallel matrix operations in one shot, and yields better results.
We demonstrate  that our \method %
outperforms %
most 
state-of-the-art instance segmentation methods in both speed and accuracy.
A light-weight version of \method  executes at 31.3 FPS and yields 37.1\%  AP on COCO \texttt{test-dev}.
Moreover, our state-of-the-art results in object detection (from our mask byproduct) and panoptic segmentation show the potential of \method to serve as a new strong baseline for many instance-level recognition tasks.
    Code is available at  \url{https://git.io/AdelaiDet}

\end{abstract}

\section{Introduction}

Generic object detection demands for the functions of localizing individual objects and recognizing their categories.
For representing the object locations, bounding box stands out for its simplicity.
Localizing objects using bounding boxes have been 
extensively
explored, including the problem formulation, network architecture, post-processing and all those focusing on optimizing and processing the bounding boxes.
The tailored solutions largely boost the performance and efficiency, thus enabling wide downstream applications recently.
However, bounding boxes  are coarse and unnatural. 
Human vision can effortlessly localize objects by their  boundaries. 
Instance segmentation, \ie, localizing objects using masks, pushes object localization to the limit at pixel level and opens up opportunities to more instance-level perception and  applications.
To date, 
most 
existing methods 
deal with instance segmentation in the view of bounding boxes, \ie, segmenting objects in (anchor) bounding boxes.
How to develop pure instance segmentation including the supporting facilities, \eg, post-processing, is largely unexplored compared to bounding box detection and instance segmentation methods built on top it. 

\begin{figure}[t!]
\begin{center}
    \includegraphics[width=1.0001\linewidth]{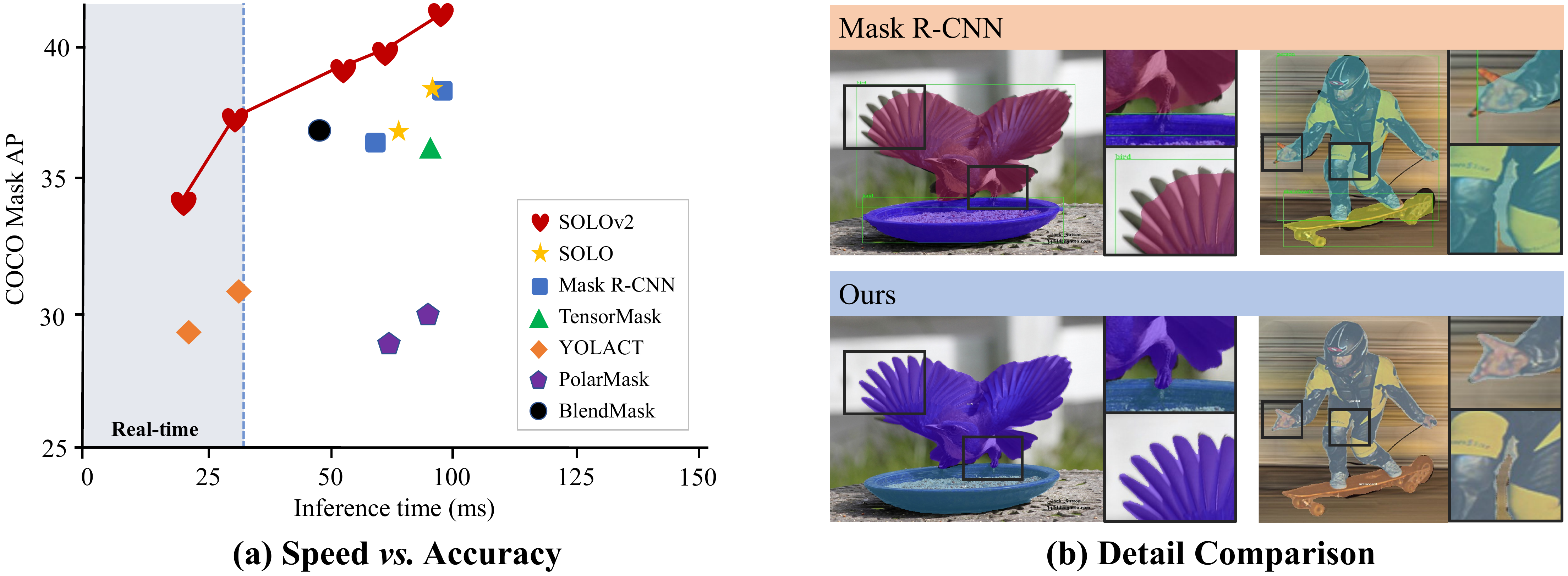}
\end{center}
\caption{
   (a) \textbf{Speed vs.\  Accuracy} on the COCO \texttt{test}-\texttt{dev}. 
   The proposed \Ours outperforms a range of state-of-the-art algorithms. Inference time of all methods is tested using one Tesla V100 GPU. 
   (b) \textbf{Detail Comparison.} \Ours depicts higher-quality masks compared with Mask R-CNN. Mask R-CNN's mask head is typically restricted to $28 \times 28$ resolution, leading to inferior prediction at object boundaries.}
\label{fig:performance}
\end{figure}

We are motivated %
by
the recently proposed SOLO (Segmenting Objects by LOcations)~\cite{solo}. 
The task of instance segmentation can be formulated as two sub-tasks of pixel-level classification, solvable using standard FCNs, thus dramatically simplifying the formulation of instance segmentation.
It takes an image as input, directly outputs instance masks and corresponding class probabilities, in a fully convolutional, box-free and grouping-free paradigm.
However, %
three main bottlenecks %
limit its performance: a) inefficient mask representation and learning; b) not high enough resolution for finer mask predictions;  c) slow mask NMS.
In this work, we eliminate the above bottlenecks all at once.
We first introduce a dynamic scheme, which enables dynamically segmenting objects by locations.
Specifically, the mask learning 
process 
can be divided into two parts: convolution kernel learning and feature learning (Figure~\ref{fig:decoupled_head_3}).
When classifying the pixels into different location categories, the 
mask kernels
are predicted 
dynamically
by the network and conditioned on the input.
We further construct a unified and high-resolution mask feature representation for instance-aware segmentation. 
As such, we are able to effortless predict high-resolution object masks, as well as learning the mask kernels and mask features separately and efficiently.

We further propose an efficient and effective matrix NMS algorithm.
As a post-processing step for suppressing the duplicate predictions, non-maximum suppression (NMS) serves as an integral part in state-of-the-art object detection systems.
Take the widely adopted multi-class NMS for example.
For each class, the predictions are sorted in descending order by confidence. 
Then for each prediction, it removes all other highly overlapped predictions. 
The sequential and recursive operations result in non-negligible latency.
For mask NMS, this drawback is further magnified.
Compared to bounding box, it takes more time to compute the IoU of each mask pair, thus leading to a large overhead.
We address this problem by introducing Matrix NMS, which performs NMS with parallel matrix operations in one shot.
Our Matrix NMS outperforms the existing NMS and its varieties in both accuracy and speed.
As a result, \textit{Matrix NMS processes 500 masks in less than 1 ms in simple python implementation}, and outperforms the recently proposed Fast NMS~\cite{yolact} by 0.4\%  AP.

With these improvements, \Ours outperforms SOLO by 1.9\%  AP while being 33\% faster. 
The Res-50-FPN \Ours  achieves 38.8\% mask AP at 18 FPS on the challenging MS COCO  dataset, evaluated on a single V100 GPU card.
A light-weight version of \Ours 
executes 
at 31.3 FPS and yields 37.1\%  mask AP.
Interestingly, although the concept of bounding box is thoroughly eliminated in our method, our bounding box byproduct, \ie, by directly converting the predicted mask to its bounding box, yield 44.9\%  AP for 
object detection, which even \textit{surpasses many state-of-the-art, highly-engineered 
object detection methods}.

We believe that, with our simple, fast and 
sufficiently strong solutions, instance segmentation should be a popular alternative to the widely used object bounding box detection, and \Ours\   may play an important role and predict its wide applications.

\subsection{Related Work}
\paragraph{Instance Segmentation}
Instance segmentation is a challenging task, as it requires instance-level and pixel-level predictions simultaneously.
The existing approaches can be summarized into three categories.
Top-down methods~\cite{fcis,maskrcnn,panet,maskscoringrcnn,Chen_2019_ICCV,yolact,chen2020blendmask,MEInst}
solve the problem from the perspective of object detection, \ie, detecting first and then segmenting the object in the box.
In particular, recent methods of \cite{chen2020blendmask,MEInst,polarmask} build their methods on 
the anchor-free object detectors~\cite{fcos}, showing promising  performance.
Bottom-up methods~\cite{associativeembedding,de2017semantic,SGN17,Gao_2019_ICCV} view the task as a label-then-cluster problem, \eg, learning the per-pixel embeddings and then clustering them into groups.
The latest direct method (SOLO)~\cite{solo} aims at dealing with instance segmentation directly, without 
dependence on 
box detection or embedding learning.
In this work, we 
appreciate
the 
basic concept
of SOLO and further explore the direct instance segmentation solutions.

We specifically compare our method with the recent YOLACT~\cite{yolact}.
YOLACT learns a group of coefficients which are normalized to (-1, 1) for each anchor box. During the inference, it first performs a bounding box detection and then uses the predicted boxes to crop the assembled masks.
While our method is evolved from SOLO~\cite{solo} through directly decoupling the original mask prediction
to kernel learning
and 
feature learning.
\textit{No anchor box is needed. No normalization is needed. No bounding box detection is needed.}
We directly map the input image to the desired object classes and object masks.
Both the training and inference are much simpler.
As a result, 
our proposed framework is much simpler, yet achieving 
significantly 
better performance (6\% AP better at 
a comparable speed); and our best model achieves 41.7\% AP vs.\ YOLACT's best 31.2\%  AP.

\paragraph{Dynamic Convolutions}
In traditional convolution layers, the learned convolution kernels stay fixed and are independent on the input, \ie, the weights are the same for arbitrary image and any location of the image.
Some previous works explore the idea of bringing more flexibility into the traditional convolutions.
Spatial Transform Networks~\cite{jaderberg2015spatial} predicts a global parametric transformation to warp the feature map, allowing the network to adaptively transform feature maps conditioned on the input.
Dynamic filter~\cite{jia2016dynamic} is proposed to actively predict the parameters of the convolution filters. It applies dynamically generated filters to an image in a sample-specific way.
Deformable Convolutional Networks~\cite{dai2017deformable}  dynamically learn the sampling locations by  predicting the offsets for each image location.
We bring the dynamic scheme into instance segmentation and enable learning instance segmenters by locations.
Yang~\etal~\cite{YangWXYK18} apply conditional \textit{batch normalization} to video object segmentation and AdaptIS~\cite{adaptis} predicts the affine parameters, which scale and shift the features conditioned on each instance.
They both belong to the more general scale-and-shift operation, which 
can  roughly be seen as an attention mechanism on intermediate feature maps.
Note that the concurrent work in \cite{CondInst} also applies dynamic convolutions for instance segmentation by extending the framework of BlendMask \cite{chen2020blendmask}.  
The dynamic scheme part is somewhat similar, but the methodology is different.
CondInst \cite{CondInst} relies on the relative position to distinguish instances as in AdaptIS, while \OurMethod uses absolute positions as in SOLO. It means that it needs to encode the position information $N$ times for $N$ instances, while \OurMethod performs it all at once using the global coordinates, 
regardless how many instances there are. 
CondInst \cite{CondInst} needs to predict at least a proposal for each instance during inference.

\paragraph{Non-Maximum Suppression}
NMS is widely adopted in many computer vision tasks and becomes an essential component of object detection 
and instance segmentation 
systems.
Some recent works~\cite{bodla2017soft, liu2019adaptive, he2019bounding, CaiZWLFAC19, yolact} are proposed to improve the traditional NMS.
They can be divided into two groups, either for improving the accuracy or speeding up.
Instead of applying the hard removal to duplicate predictions according to a threshold,  Soft-NMS~\cite{bodla2017soft} decreases the confidence scores of neighbors according to their overlap with higher scored predictions.
Adaptive NMS~\cite{liu2019adaptive} applies dynamic suppression threshold to each instance, which is tailored for pedestrian detection in a crowd.
In~\cite{he2019bounding}, the authors use KL-Divergence and reflected it in the refinement of coordinates in the NMS process.
To accelerate the inference, Fast NMS~\cite{yolact} enables deciding the predictions to be kept or discarded in parallel. 
Note that 
it speeds up at the cost of performance
deterioration. 
Different from the previous methods, our Matrix NMS addresses the issues of hard removal and sequential operations at the same time.
As a result, \textit{the proposed Matrix NMS is able to process 500 masks in less than 1 ms} in simple python implementation, which is 
negligible compared with the time of network
evaluation, 
and yields 0.4\%  AP better than Fast NMS.

\section{Our Method: \method} 
\begin{figure}[t]
\centering
\subfigure[SOLO]{
\includegraphics[width=0.6058746\textwidth]{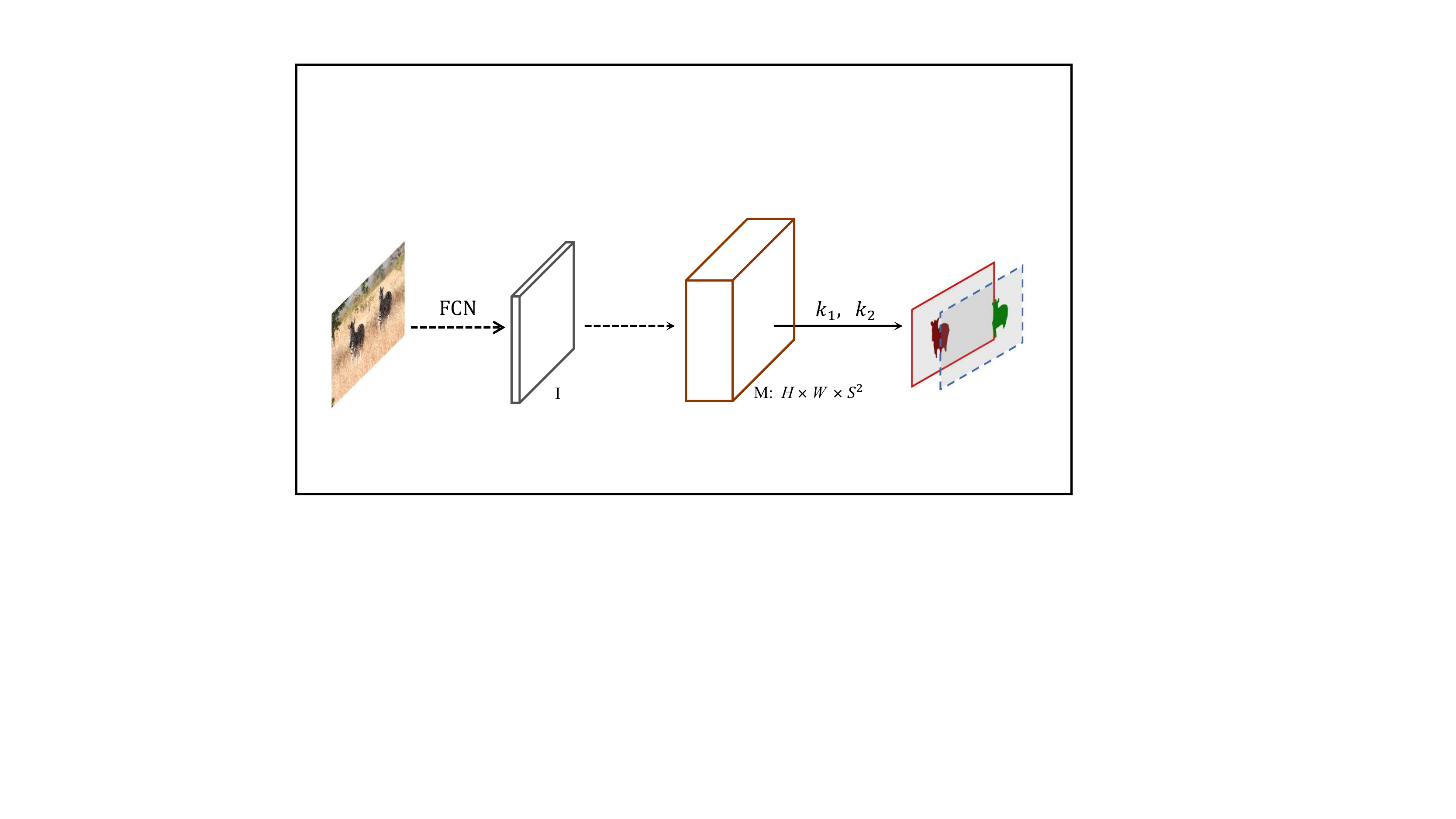}
\label{fig:decoupled_head_1}
}
\hspace{0.16in}
\subfigure[\OurMethod]{
\includegraphics[width=0.6058746\textwidth]{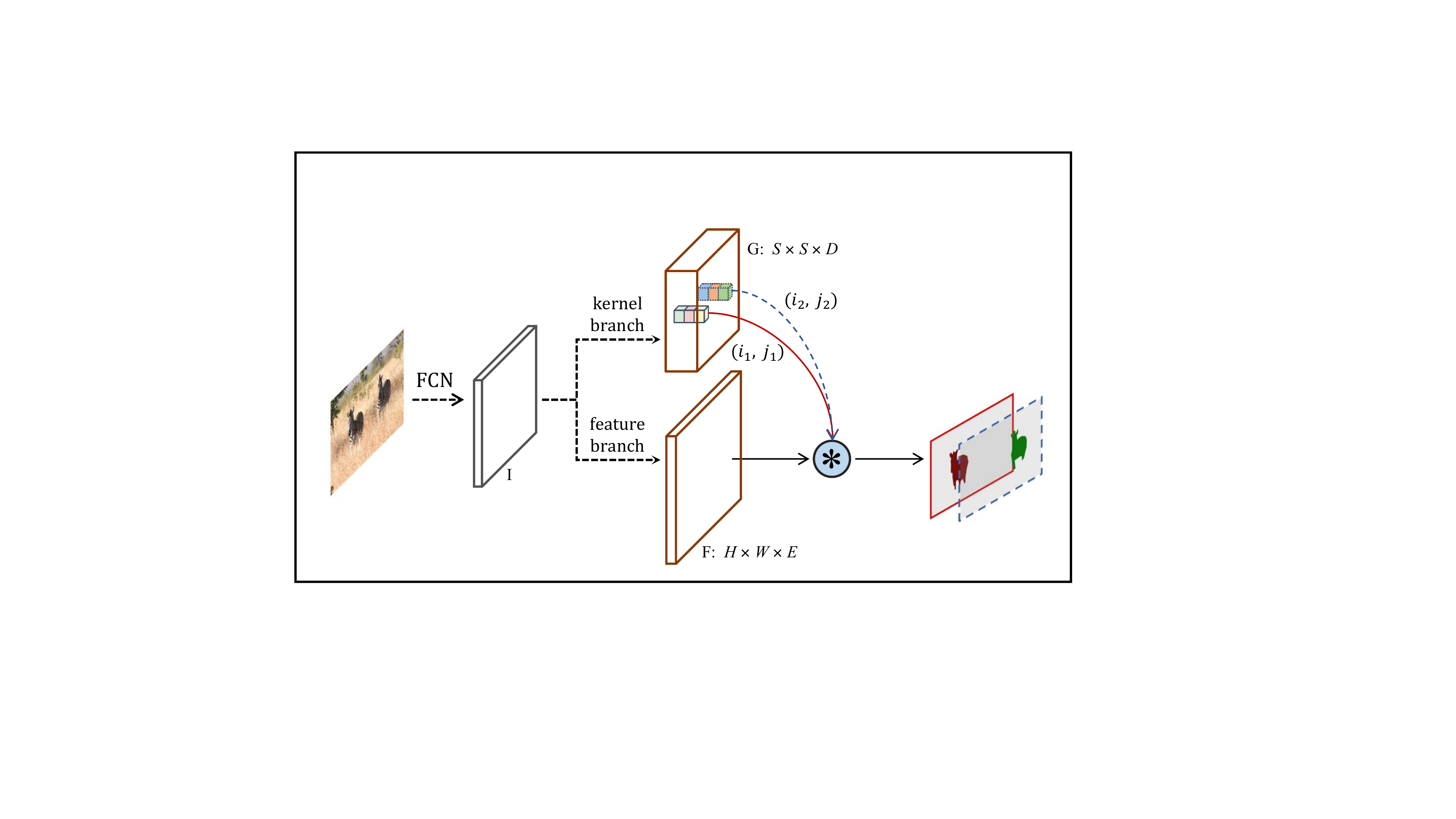}
\label{fig:decoupled_head_3}
}
\caption{\textbf{\OurMethod} %
compared to SOLO. $I$ is the input feature after FCN-backbone representation extraction. Dashed arrows denote convolutions. $k=i \cdot  S + j$; and  
`$\circledast$' denotes the dynamic convolution operation.}
\label{fig:dynamic_head}
\end{figure}

An instance segmentation system should separate different instances at pixel level. 
To distinguish instances, we follow the basic concept of `segmenting  objects by locations'~\cite{solo}. 
The input image is conceptually divided into $S \times S$ grids. 
If the center of an object falls into a grid cell, then the grid cell corresponds to a binary mask for that object. 
As such, the system outputs $S^2$ masks in total, denoted as $M\in \mathbb{ R}^{H\times W\times S^2}$. 
The $k^{th}$ channel is responsible for segmenting instance at position ($i$, $j$), where $k=i\cdot S + j$ (see Figure~\ref{fig:decoupled_head_1}).

Such paradigm could generate the instance segmentation results in an elegant way. 
However, there are three main bottlenecks that limit its performance: 
a) inefficient mask representation and learning. 
It takes a lot of memory and computation to predict the output tensor $M$, which has $S^2$ channels. 
Besides, as the $S$ is different for different FPN level, the last layer of each level is learned separately and not shared, which results in an inefficient training.
b) inaccurate mask predictions. 
Finer predictions require high-resolution masks to deal with the details at object boundaries. 
But large resolutions will considerably increase the computational cost.
c) slow mask NMS. Compared with box NMS, mask NMS takes more time and leads to a larger overhead.

In this section, we show that these challenges can be effectively solved by our proposed dynamic mask representation and Matrix NMS, and we introduce them as follows.

\subsection{Dynamic Instance Segmentation}
We first revisit the mask generation in SOLO~\cite{solo}.
To generate the instance mask of $S^2$ channels corresponding to $S\times S$ grids, the last layer takes one level of pyramid features $F\in \mathbb{ R}^{H\times W\times E}$ as input and at last applies a convolution layer with $S^2$ output channels.
The operation can be written as:
\begin{equation}
\label{eq:convop}
\begin{aligned}
M_{i,j} = G_{i,j} * F,
\end{aligned}
\end{equation}
where $G_{i,j}\in \mathbb{ R}^{1\times 1\times E}$ is the conv kernel, and $M_{i,j}\in \mathbb{R}^{H\times W}$ is the final mask containing only one instance whose center is at location $(i,j)$.

In other words, we need two input $F$ and $G$ to generate the final mask $M$.
Previous work explicitly output the whole $M$ for training and inference. 
Note that tensor $M$ is very large, and to directly predict $M$ is memory and computational inefficient. 
In most cases the objects are located sparsely in the image.
$M$ is redundant as only a small part of $S^2$ kernels actually functions during a single inference.

From another perspective, if we separately learn $F$ and $G$, the final $M$ could be directly generated using the both components.
In this way, we can simply pick the valid ones from predicted $S^2$ kernels and perform the convolution dynamically.
The number of model parameters also decreases. 
What's more, as the predicted kernel is generated dynamically conditioned on the input, it benefits from the flexibility and adaptive nature.
Additionally, each of $S^2$ kernels is conditioned on the location.
It is in accordance with the core idea of segmenting objects by locations and goes a step further by predicting the segmenters by locations.

\subsubsection{Mask Kernel $G$}
Given the backbone and FPN, we predict the mask kernel $G$ at each pyramid level. 
We first resize the input feature $F_I \in \mathbb{ R}^{H_I\times W_I\times C}$ into shape of $S\times S\times C$. 
Then 4\X convs and a final $3\times 3\times D$ conv are employed to generate the kernel $G$. 
We add the spatial functionality to $F_I$ by giving the first convolution access to the normalized coordinates following CoordConv~\cite{coordconv}, \ie, concatenating two additional input channels which contains pixel coordinates normalized to $[-1, 1]$. Weights for the head are shared
across different feature map levels.

For each grid, the kernel branch predicts the $D$-dimensional output to indicate predicted convolution kernel weights, where $D$ is the number of parameters.
For generating the weights of a $1$$\times$$1$ convolution with $E$ input channels, $D$ equals $E$.
As for $3$$\times$$3$ convolution, $D$ equals $9E$.
These generated weights are conditioned on the locations, \ie, the grid cells. 
If we divide the input image
into $S$$\times$$S$ grids, the output space will be $S$$\times$$S$$\times$$D$, There is no activation function on the output.

\subsubsection{Mask Feature $F$}

Since the mask feature and mask kernel are decoupled and separately predicted, there are two ways to construct the mask feature.
We can put it into the head, along with the kernel branch. It means that we predict the mask features for each FPN level.
Or, to predict a unified mask feature representation for all FPN levels.
We have compared the two implementations in Section~\ref{subsubsec:ablation} by experiments.
Finally, we employ the latter one for its effectiveness and efficiency.

For learning a 
unified and high-resolution mask feature representation, we apply  feature pyramid fusion inspired by the semantic segmentation in~\cite{kirillov2019panoptic}.
After repeated stages of $3\times3$ conv, group norm~\cite{wu2018group}, ReLU and $2\times$ bilinear upsampling, the FPN features P2 to P5 are merged into a single output at 1/4 scale.
The last layer after the element-wise summation consists of $1 \times 1$ convolution, group norm and ReLU.
More details can be referred to supplementary material.
It should be noted that we
feed normalized pixel coordinates to the deepest FPN level (at
1/32 scale), before the convolutions and bilinear upsamplings. 
The provided accurate position information is important for enabling position sensitivity and predicting instance-aware features.

\subsubsection{Forming Instance Mask} For each grid cell at ($i, j$), we first obtain the mask kernel $G_{i, j, :} \in {\mathbb  R}^{D}$. Then $G_{i, j, :}$ is convolved with $F$ to get the instance mask. In total, there will be at most $S^2$ masks for each prediction level. Finally, we use the proposed Matrix NMS to get the final instance segmentation results.

\subsubsection{Learning and Inference}
The training loss function is defined as follows: 
\begin{equation}
\label{eq:loss_all}
\begin{aligned}
L = L_{cate} + \lambda L_{mask}, 
\end{aligned}
\end{equation}
where $L_{cate}$ is the conventional Focal Loss~\cite{focalloss} for semantic category classification, $L_{mask}$ is the Dice Loss for mask prediction. 
For more details, we refer readers to~\cite{solo}.

During the inference, 
we forward input image through the backbone network and FPN, and obtain the category score
$\mathbf{p}_{i,j}$ at grid $(i, j)$.
We first use a confidence threshold of $0.1$ to filter out predictions with low confidence.
The corresponding predicted mask kernels are then used to perform convolution on the mask feature.
After the \texttt{sigmoid} operation, we use a threshold of $0.5$ to convert predicted soft masks to binary masks.
The last step is the Matrix NMS.

\def\iou{{{\tt iou}}}
\subsection{Matrix NMS}

\myparagraph{Motivation}
Our Matrix NMS is motivated by Soft-NMS~\cite{bodla2017soft}. Soft-NMS decays the other detection scores as a monotonic decreasing function $f(\iou)$ of their overlaps. By decaying the scores according to IoUs recursively, higher IoU detections will be eliminated with a minimum score threshold. However, such process is sequential like traditional Greedy NMS and could not be implemented in parallel.

Matrix NMS views this process from another perspective by considering how a predicted mask $m_j$ being suppressed. For $m_j$, its decay factor is affected by: (a) The penalty of each prediction $m_{i}$ on $m_j$ ($s_i>s_j$), where $s_i$ and $s_j$ are the confidence scores; and (b) the probability of $m_i$ being suppressed. For (a), the penalty of each prediction $m_{i}$ on $m_j$ could be easily computed by $f(\iou_{i,j})$. For (b), the probability of $m_i$ being suppressed is not so elegant to be computed. However, the probability usually has positive correlation with the IoUs. So here we directly approximate the probability by the most overlapped prediction on $m_i$ as 
\begin{equation}
\label{eq:prob}
    f(\iou_{\cdot, i}) = \min_{\forall s_k>s_i}f(\iou_{k, i}).
\end{equation}
To this end, the final decay factor becomes
\begin{equation}
\label{eq:matrixnms}
    decay_j = \min_{\forall s_i>s_j} \frac{f(\iou_{i, j})}{f(\iou_{\cdot, i})},
\end{equation}
and the updated score is computed  by $s_j = s_j \cdot decay_j$.
We consider two most simple decremented functions, denoted as \texttt{linear} %
$
    f(\iou_{i,j}) = 1 - \iou_{i,j},
$
and \texttt{Gaussian}
$
    f(\iou_{i,j})= \exp \Big( {-\frac{\iou_{i,j}^2}{\sigma}}
                       \Big).
$

\myparagraph{Implementation} All the operations in Matrix NMS could be implemented in one shot without recurrence. We first compute a $N\times N$ pairwise IoU matrix for the top $N$ predictions sorted descending by score. For binary masks, the IoU matrix could be efficiently implemented by matrix operations. Then we get the most overlapping IoUs by column-wise max on the IoU matrix. Next, the decay factors of all higher scoring predictions are computed, and the decay factor for each prediction is selected as the most effect one by column-wise min (Eqn.~\eqref{eq:matrixnms}). Finally, the scores are updated by the decay factors. For usage, we just need thresholding and selecting top-$k$ scoring masks as the final predictions.

The pseudo-code of Matrix NMS is provided in supplementary material. In our code base, Matrix NMS is 9\X  faster than traditional NMS and being more accurate (Table~\ref{tab:nms}(c)). 
We show that Matrix NMS serves as a superior alternative of traditional NMS both in accuracy and speed, and can be easily integrated into the state-of-the-art detection/segmentation systems.

\begin{table*}[t]
\centering
\footnotesize 
\caption{\textbf{Instance segmentation} mask AP (\%) on COCO \texttt{test}-\texttt{dev}. All entries are \textit{single-model} results. Mask R-CNN$ ^*$ is our improved version with scale augmentation and longer training time ($6\times$).
`DCN' means deformable convolutions used.
}
\begin{tabular}{ r | l |cccccc}
\footnotesize
& backbone &AP & AP$_{50}$ & AP$_{75}$&AP$_{S}$ & AP$_{M}$ & AP$_{L}$\\
\Xhline{1pt}
\emph{box-based:} &&&&&&&\\
~~~~Mask R-CNN~\cite{maskrcnn}        &Res-101-FPN &35.7  &58.0   &37.8   &15.5   &38.1   &52.4 \\
~~~~Mask R-CNN$^*$                    &Res-101-FPN &37.8  &59.8   &40.7   &\textbf{20.5}   &40.4   &49.3 \\
~~~~MaskLab+~\cite{masklab}           &Res-101-C4  &37.3  & 59.8   & 39.6   & 16.9   & 39.9   &53.5 \\
~~~~TensorMask~\cite{Chen_2019_ICCV}  &Res-101-FPN &37.1  &59.3   &39.4   &17.4   &39.1   &51.6 \\
~~~~YOLACT~\cite{yolact}              &Res-101-FPN &31.2  &50.6   &32.8   &12.1   &33.3   &47.1 \\
~~~~MEInst~\cite{MEInst}              &Res-101-FPN &33.9  &56.2   &35.4   &19.8   &36.1   &42.3  \\
~~~~CenterMask~\cite{wang2020centermask} &Hourglass-104 &34.5 &56.1  &36.3   &16.3   &37.4   &48.4  \\
~~~~BlendMask~\cite{chen2020blendmask} &Res-101-FPN &38.4  &60.7  &41.3    &18.2   &41.5   &53.3 \\
\hline\hline
\emph{box-free:} &&&&&&&\\
~~~~PolarMask~\cite{polarmask}        &Res-101-FPN &32.1 &53.7 &33.1 &14.7 &33.8 &45.3 \\
~~~~SOLO~\cite{solo}      &Res-101-FPN &37.8 & 59.5 & 40.4 & 16.4 & 40.6 & 54.2\\
~~~~\textbf{\OurMethod}               &Res-50-FPN  &  38.8 & 59.9  & 41.7  & 16.5  & 41.7  & 56.2 \\
~~~~\textbf{\OurMethod}               &Res-101-FPN  &  39.7 & 60.7 & 42.9 & 17.3 & 42.9 & 57.4 \\
~~~~\textbf{\OurMethod}               &Res-DCN-101-FPN  & \textbf{41.7} & \textbf{63.2} & \textbf{45.1} & 18.0 & \textbf{45.0} & \textbf{61.6} \\
\end{tabular}
\label{tab:sota}
\end{table*}

\section{Experiments}
To evaluate the proposed method \Ours, we conduct experiments on three basic tasks, instance segmentation, object detection, and panoptic segmentation on MS COCO~\cite{coco}. We also present experimental results on the recently proposed LVIS dataset~\cite{lvis2019}, which has more than 1K categories and thus is 
considerably more challenging.

\subsection{Instance segmentation}
For instance segmentation, we report lesion and sensitivity studies by evaluating on the COCO 5K \texttt{val2017} split. We also report COCO mask AP on the
\texttt{test}-\texttt{dev} split, which 
is evaluated on the evaluation server. 
\method is trained with stochastic gradient descent (SGD). We use synchronized SGD over 8 GPUs with a total of 16 images per mini-batch.
Unless otherwise specified, all models are trained for  36 epochs (\ie, 3$ \times$) with an initial learning rate of $0.01$, which is then
divided by 10 at 27th and again at 33th epoch. 
We use scale jitter where the shorter
image side is randomly sampled from 640 to 800 pixels.

\subsubsection{Main Results}

We compare \OurMethod to the state-of-the-art methods in instance segmentation on MS COCO \texttt{test}-\texttt{dev} in Table~\ref{tab:sota}. \OurMethod with ResNet-101 achieves a mask AP of 39.7\%, which is much better than 
other state-of-the-art instance segmentation methods. 
Our method shows its superiority especially on large objects (\eg, ~+5.0 AP$_{L}$ than Mask R-CNN). 

We also provide the speed-accuracy trade-off on COCO to compare with some dominant instance segmenters (Figure~\ref{fig:performance} (a)). 
We show our models with ResNet-50, ResNet-101, ResNet-DCN-101 and two light-weight versions described in Section~\ref{subsubsec:ablation}.
The proposed \OurMethod outperforms a range of state-of-the-art algorithms, both in accuracy and speed.  The running time is tested on our local machine, with a single V100 GPU.
We download code and pre-trained models to test inference time for each model on the same machine.
Further, as described in Figure~\ref{fig:performance} (b), \OurMethod predicts much finer masks than Mask R-CNN which performs on the local region. 

Beside the MS COCO dataset, we also demonstrate the effectiveness of \OurMethod on LVIS dataset.
Table~\ref{table:lvis} 
reports the performances on the 
rare (1$\sim$10 images), common (11$\sim$100), and frequent ($>$ 100) subsets, as well as the overall AP. 
Both the reported Mask R-CNN and \Ours use data resampling training strategy, following~\cite{lvis2019}.
Our \Ours outperforms the baseline method by about 1\% AP. For large-size  objects (AP$_L$), our \Ours 
achieves 6.7\%  AP improvement, which is consistent with the results on the COCO dataset.

\begin{table*}[t]
\centering
\caption{Instance segmentation results on the 
{LVIS}v0.5 validation dataset. $*$ means re-implementation.}
\vspace{0.1in}
\small
\begin{tabular}{ r  |l|ccc|ccc|c}
  & backbone & AP$_r$ & AP$_c$ & AP$_f$ & AP$_S$ & AP$_M$ & AP$_L$&AP \\
\Xhline{1pt}
Mask-RCNN~\cite{lvis2019}& Res-50-FPN     & 14.5 & 24.3 & 28.4 & - & - & - & 24.4 \\
Mask-RCNN$^*$-3\X & Res-50-FPN               & 12.1 & 25.8 & 28.1 & \textbf{18.7} & 31.2 & 38.2 & 24.6 \\
\hline
\textbf{\Ours}  & Res-50-FPN      & 13.4 & 26.6 & 28.9 & 15.9 & 34.6 & 44.9 & 25.5 \\
\textbf{\Ours}  & Res-101-FPN     & \textbf{16.3} & \textbf{27.6} & \textbf{30.1} & 16.8 & \textbf{35.8} & \textbf{47.0} & \textbf{26.8}
\end{tabular}
\label{table:lvis}
\end{table*}

\begin{table}[!t]
\caption{Ablation experiments for \OurMethod. All models are trained on MS COCO \texttt{train2017}, test on \texttt{val2017} unless noted.}
\vskip 0.05 in
\begin{minipage}{0.3\linewidth}
\begingroup
\footnotesize{ (a) \textbf{Kernel shape.}  The performance is stable when the shape goes beyond $1\times1\times256$.}
\endgroup
\vskip 0.05 in
\centering
\scalebox{0.7}{
\begin{tabular}{c|ccc}
        Kernel shape &AP & AP$_{50}$ & AP$_{75}$\\
        \Xhline{1pt}
        $3\times3\times64$ &  37.4 & 58.0 & 39.9 \\
        $1\times1\times64$&  37.4  & 58.1  & 40.1  \\
        $1\times1\times128$&  37.4 & 58.1 & 40.2   \\
        $1\times1\times256$&  \textbf{37.8} & \textbf{58.5} & \textbf{40.4}   \\
        $1\times1\times512$&  37.7 & 58.3 & \textbf{40.4}    \\
    \end{tabular}}
\label{tab:kernel_shape}
\end{minipage}
\hfill
\begin{minipage}{.32\linewidth}
\begingroup
\footnotesize{ (b) \textbf{Explicit coordinates.} Precise coordinates input can considerably improve the results.}
\endgroup
\vskip 0.05 in
\centering
\scalebox{0.75}{
\begin{tabular}{p{.7cm}<{\centering} p{1cm}<{\centering} |ccc}
        Kernel & Feature &AP & AP$_{50}$ & AP$_{75}$ \\
        \Xhline{1pt}
         & & 36.3 & 57.4 & 38.6     \\
         \cmark & &  36.3  & 57.3  & 38.5   \\
        & \cmark &  37.1 & 58.0 & 39.4   \\
        \cmark & \cmark & \textbf{37.8} & \textbf{58.5} & \textbf{40.4}  \\
    \end{tabular}
}
\label{tab:coordinates}
\end{minipage}
\hfill
\begin{minipage}{.32\linewidth}
\begingroup
\footnotesize{ (c) \textbf{Matrix NMS.} Matrix NMS outperforms other methods in both speed and accuracy.}
\endgroup
\vskip 0.05 in
\centering
\scalebox{0.7}{
\begin{tabular}{ r |ccc}
        Method & Iter? & Time(ms) & AP \\
        \Xhline{1pt}
         Hard-NMS & \cmark & 9 & 36.3    \\
         Soft-NMS & \cmark  & 22   & 36.5   \\
          Fast NMS & \xmark & $\textless1$ &  36.2  \\
         Matrix NMS & \xmark & $\textless1$ & \textbf{36.6}   \\
    \end{tabular}
}
\label{tab:nms}
\end{minipage}
\vskip 0.2 in
\begin{minipage}{.32\linewidth}
\begingroup
\footnotesize{ (d) \textbf{Mask feature representation.} We compare the separate mask feature representation in parallel heads and the unified representation.}
\endgroup
\vskip 0.05 in
\centering
\scalebox{0.75}{
\begin{tabular}{c|ccc}
        Mask Feature &AP & AP$_{50}$ & AP$_{75}$\\
        \Xhline{1pt}
        Separate  &  37.3 & 58.2 & 40.0\\
        Unified &  \textbf{37.8} & \textbf{58.5} & \textbf{40.4}  \\
    \end{tabular}
}
\label{tab:mask_fea}
\end{minipage}
\hfill
\begin{minipage}{.32\linewidth}
\begingroup
\footnotesize{ (e) \textbf{Training schedule.} 1$\times$ means 12 epochs using single-scale training. 3$\times$ means 36 epochs with multi-scale training.}
\endgroup
\vskip 0.05 in
\centering
\scalebox{0.75}{
\begin{tabular}{c|ccc}
        Schedule &AP & AP$_{50}$ & AP$_{75}$ \\
        \Xhline{1pt}
        1$\times$ &  34.8 & 54.8 & 36.8   \\
        3$\times$ &  \textbf{37.8} & \textbf{58.5} & \textbf{40.4}  \\
    \end{tabular}
}
\label{tab:dynamic_head}
\end{minipage}
\hfill
\begin{minipage}{.32\linewidth}
\begingroup
\footnotesize{ (f) \textbf{Real-time \OurMethod}. The speed is reported on a single V100 GPU by averaging 5 runs (on COCO \texttt{test-dev}). }
\endgroup
\vskip 0.05 in
\centering
\scalebox{0.68}{
\begin{tabular}{c|c c c c}
    Model  & AP & AP$_{50}$ & AP$_{75}$ & fps \\
        \Xhline{1pt}
         \method-448  & 34.0 & 54.0 & 36.1 & 46.5   \\
         \method-512  & 37.1 & 57.7 & 39.7 & 31.3
    \end{tabular}
}
\label{tab:real_time}
\end{minipage}
\label{table:ablation}
\end{table}

\subsubsection{Ablation Experiments}
\label{subsubsec:ablation}

We investigate and compare the following five aspects in our methods.

\myparagraph{Kernel shape}
We consider the kernel shape from two aspects: number of input channels and kernel size.
The comparisons are shown in Table~\ref{tab:kernel_shape}(a).
$1\times1$ conv shows equivalent performance to $3\times3$ conv.
Changing the number of input channels from 128 to 256 
attains
0.4\%  AP gains.  When it grows beyond 256, the performance becomes stable. In this work, we  set the number of input channels to be 256 in all other experiments.

\myparagraph{Effectiveness of coordinates}
Since our method segments objects by locations, or specifically, learns the object segmenters by locations, the position information is very important.
For example, if the mask kernel branch is unaware of the positions, the objects with the same appearance may have the same predicted kernel, leading to the same output mask.
On the other hand, if the mask feature branch is unaware of the position information, it would not know how to assign the pixels to different feature channels in the order that matches the mask kernel.
As shown in Table~\ref{tab:coordinates}(b),
the model achieves 36.3\%  AP without explicit coordinates input.
The results are 
reasonably good 
because that CNNs can implicitly learn the absolute position information from the commonly used zero-padding operation, as revealed in~\cite{Islam2020How}.
The pyramid zero-paddings in our mask feature branch should have contributed considerably. 
However, the implicitly learned position information is coarse and inaccurate.
When making the convolution access to its own input coordinates through 
concatenating extra coordinate channels,
our method enjoys 1.5\% absolute AP gains.

\myparagraph{Unified Mask Feature Representation}
For mask feature learning, we have two options: to learn the feature in the head separately for each FPN level or to construct a unified representation.
For the former one, we implement as SOLO and use 
seven 
$3\times3$ conv to predict the mask features.
For the latter one, we fuse the FPN's features in a simple way and obtain the unified mask representations. The detailed implementation is in supplementary material.
We compare these two modes in Table~\ref{tab:mask_fea}(d).
As shown, the unified representation achieves better results, especially for the medium and large objects.
This is easy to understand: 
In separate way, the large-size objects are assigned to 
high-level feature maps of low spatial resolutions, leading to 
coarse boundary prediction.

\iffalse
\myparagraph{Dynamic vs.\ Decoupled}
 The dynamic head and decoupled head both serve as the efficient varieties of the SOLO head.
 We compare the results in Table~\ref{tab:dynamic_head}(e).
 All the settings are the same except the head type, which means that for the dynamic head we use the separate features as above.
 The  dynamic head 
 %
 achieves
 0.7\% AP better than the decoupled head.
We believe that the gains 
%
have
come from the dynamic scheme which learns the kernel weights dynamically,
conditioned on the input.
\fi

\myparagraph{Matrix NMS}  Our Matrix NMS can be implemented totally in parallel. Table~\ref{tab:nms}(c) 
presents
the speed and accuracy comparison of Hard-NMS, Soft-NMS, Fast NMS and our Matrix NMS. Since all methods need to compute the IoU matrix, we pre-compute the IoU matrix in advance for fair comparison. The speed reported here is 
that of the 
NMS process alone, excluding computing IoU matrices. 
Hard-NMS and Soft-NMS are widely used in current object detection and segmentation models. Unfortunately, both methods are 
recursive
and spend much time budget (\eg 22 ms). Our Matrix NMS only needs 
$\textless 1$ ms and is almost cost free! Here we also show the performance of Fast NMS, which utilizes matrix operations but with performance penalty. To conclude, our Matrix NMS shows its advantages on both speed and accuracy.

\myparagraph{Real-time setting}
We design two light-weight models for different purposes. 
1) \texttt{Speed priority}, the number of 
convolution layers in the prediction head is reduced to two and the input shorter side is 448.
2) \texttt{Accuracy priority}, the number of 
convolution layers in the prediction head is reduced to three and the input shorter side is 512. Moreover, deformable convolution~\cite{dai2017deformable} is used in the backbone and the last layer of prediction head. 
We train both models with the $3\times$ schedule, with shorter side randomly sampled from [352, 512].
Results are shown in Table~\ref{tab:real_time}(f).
\OurMethod can not only push state-of-the-art, but has also been ready for real-time applications. 

\subsection{Extensions: Object Detection and Panoptic Segmentation}

\begin{figure}[t]
\begin{minipage}{0.42\linewidth}
\centering 
    \includegraphics[width=0.88\linewidth]{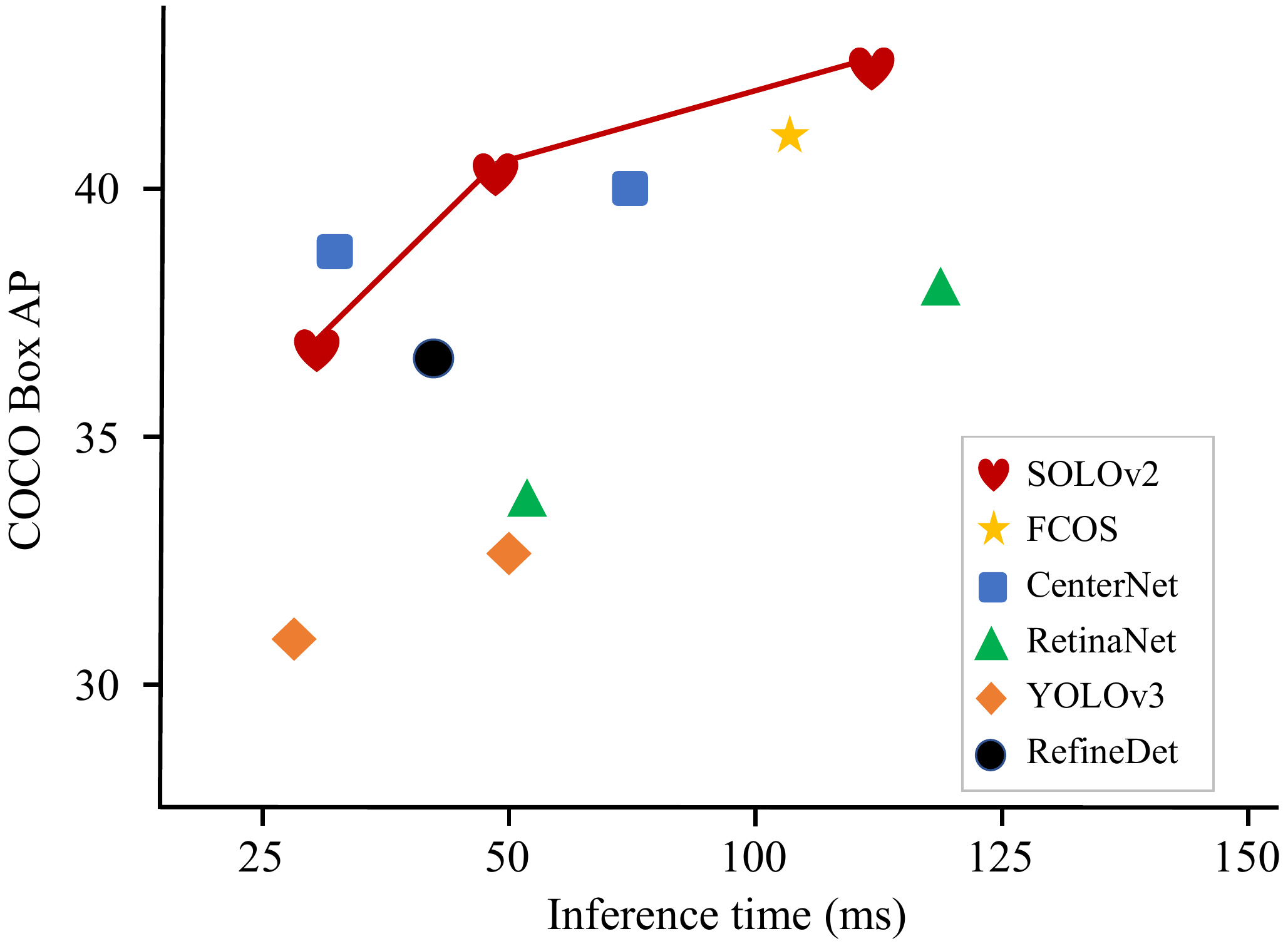}
\vskip 0.05 in
\begingroup
\footnotesize{ (a) Speed-accuracy trade-off of bounding-box \textbf{object detection} on the COCO \texttt{test}-\texttt{dev}.}
\endgroup
\label{fig:box_speed_ap}
\end{minipage}
\hspace{0.3in}
\begin{minipage}{0.42\linewidth}
\centering
\footnotesize
\scalebox{0.8}{\begin{tabular}{ r |c|cc}
  &  PQ & PQ\textsuperscript{Th} & PQ\textsuperscript{St} \\
\Xhline{1pt}
\emph{box-based:} &&&\\
~~~~AUNet~\cite{li2019attention}       &  39.6 & 49.1 & 25.2  \\
~~~~UPSNet~\cite{xiong2019upsnet}        &  \textbf{42.5} & 48.5 & \textbf{33.4} \\

~~~~Panoptic-FPN~\cite{kirillov2019panoptic} &  39.0 & 45.9 & 28.7   \\
~~~~Panoptic-FPN$^*$-1\X      &  38.7 & 45.9 & 27.8  \\
~~~~Panoptic-FPN$^*$-3\X    & 40.8 & 48.3 & 29.4 \\
\hline\hline
\emph{box-free:} &&&\\
~~~~AdaptIS~\cite{adaptis}        & 35.9 & 40.3  & 29.3   \\
~~~~SSAP ~\cite{Gao_2019_ICCV}  & 36.5 & $-$ & $-$ \\
~~~~Pano-DeepLab ~\cite{cheng2019panoptic}  & 39.7 & 43.9 & \textbf{33.2} \\
~~~~\textbf{\Ours}             & \textbf{42.1} & \textbf{49.6} & 30.7  \\
\end{tabular}}
\vskip 0.2 in
\begingroup
\footnotesize{ (b) \textbf{Panoptic segmentation} results on COCO val2017. $*$ means re-implementation.}
\endgroup
\end{minipage}
\vskip 0.1 in
\caption{Extensions on object detection and panoptic segmentation.}
\label{fig:extensions}
\end{figure}

Although our instance segmentation solution removes the dependence of bounding box prediction, we are able to produce the 4-d object bounding box from each instance mask.
The best model of ours achieve 44.9 AP on COCO {\tt test-dev}.
\OurMethod beats most recent methods in both accuracy and speed, as shown in Figure~\ref{fig:extensions}(a).
Here we emphasize that our results are directly generated from the off-the-shelf instance mask, without any box based supervised training or engineering.

Besides, we also demonstrate the effectiveness of \OurMethod on the problem of panoptic segmentation.
The proposed \Ours can be easily extended to panoptic segmentation by adding the semantic segmentation branch, analogue to the mask feature branch.
We use annotations of COCO 2018 panoptic segmentaiton task. All models are trained on {\tt train2017}
subset and tested on {\tt  val2017}. 
We use the same strategy as in Panoptic-FPN to combine instance and semantic results. 
As shown in Figure~\ref{fig:extensions}(b), our method achieves state-of-the-art results and outperforms  other recent box-free methods by a large margin. All methods listed use the same backbone (ResNet50-FPN) except SSAP (ResNet101) and Pano-DeepLab (Xception-71).
Note that UPSNet has used deformable convolution~\cite{dai2017deformable} for better performance.

\section{Conclusion}

In this work, we have introduced a dynamic and fast instance segmentation solution with strong performance, from three aspects.
\begin{itemize}
\itemsep 0pt
    \item  We have proposed to learn adaptive, dynamic convolutional 
    kernels for the mask prediction, conditioned on the location, %
    leading to a much more compact yet more powerful head design, and achieving better results.
    \item 
    We have re-designed the object mask generation in a simple and unified way, which predicts more accurate boundaries.
    \item 
    Moreover, 
    unlike box NMS as in object detection, for direct instance segmentation 
    a bottleneck in inference efficiency is the NMS of masks.
    We have designed a simple and much faster NMS strategy, termed Matrix NMS, for NMS processing of masks, without sacrificing mask AP.
\end{itemize}

Our experiments on the MS COCO and LVIS datasets demonstrate the 
superior performance in terms of both accuracy and speed of the proposed \OurMethod. 
Being versatile for instance-level recognition tasks, we show that 
without any modification to the framework, \Ours performs competitively for 
panoptic segmentation. Thanks to its simplicity (being proposal free, anchor free, 
FCN-like), strong performance in both accuracy and speed, and potentially 
being capable of solving many instance-level tasks, we hope that \Ours can be a strong baseline approach to instance recognition, and inspires future work 
such that its full potential can be exploited as we believe that
there is still much room for improvement.

\textit{
T. Kong and C. Shen are the corresponding authors.
C. Shen and  
his employer received no financial support for the research, authorship, and/or publication of this article.
}

\section*{Appendix}

\def\x{{$\times$}}
\def\X{{$\times$}}

\appendix
\section{Matrix NMS}
The pseudo-code of Matrix NNS is shown in Figure~\ref{fig:code}.
All the operations in Matrix NMS could be implemented in one shot without recurrence. 
In our code base, Matrix NMS is 9\X \ times faster than traditional NMS and being more accurate. 
We show that Matrix NMS serves as a superior alternative of traditional NMS both in accuracy and speed, and can be easily integrated into the state-of-the-art detection/segmentation systems.

% ------------------------------------------------------------------
% Algorithm Matrix NMS
\lstset{
  backgroundcolor=\color{white},
  basicstyle=\fontsize{7.5pt}{8.5pt}\fontfamily{lmtt}\selectfont,
  columns=fullflexible,
  breaklines=true,
  captionpos=b,
  commentstyle=\fontsize{8pt}{9pt}\color{codegray},
  keywordstyle=\fontsize{8pt}{9pt}\color{codegreen},
  stringstyle=\fontsize{8pt}{9pt}\color{codeblue},
  frame=tb,
  otherkeywords = {self},
}
\begin{figure}[!h]
\begin{lstlisting}[language=python]
def matrix_nms(scores, masks, method='gauss', sigma=0.5):
    # scores: mask scores in descending order (N)
    # masks: binary masks (NxHxW)
    # method: 'linear' or 'gauss'
    # sigma: std in gaussian method

    # reshape for computation: Nx(HW)
    masks = masks.reshape(N, HxW)
    # pre-compute the IoU matrix: NxN
    intersection = mm(masks, masks.T) 
    areas = masks.sum(dim=1).expand(N, N)
    union = areas + areas.T - intersection
    ious = (intersection / union).triu(diagonal=1)
    
    # max IoU for each: NxN 
    ious_cmax = ious.max(0)
    ious_cmax = ious_cmax.expand(N, N).T
    # Matrix NMS, Eqn.(4): NxN
    if method == 'gauss':     # gaussian 
        decay = exp(-(ious^2 - ious_cmax^2) / sigma)
    else:                     # linear
        decay = (1 - ious) / (1 - ious_cmax)
    # decay factor: N
    decay = decay.min(dim=0)
    return scores * decay
\end{lstlisting}
\caption{Python code of Matrix NMS. \texttt{mm}: matrix multiplication; \texttt{T}: transpose; \texttt{triu}: upper triangular part}
\label{fig:code}
\end{figure}

\section{Unified Mask Feature Representation}
The detailed implementation is illustrated in Figure~\ref{fig:mask_feature}.
For learning 
 a 
unified and high-resolution mask feature representation, we apply  feature pyramid fusion inspired by the semantic segmentation in~\cite{kirillov2019panoptic}.
After repeated stages of $3\times3$ conv, group norm~\cite{wu2018group}, ReLU and $2\times$ bilinear upsampling, the FPN features P2 to P5 are merged into a single output at 1/4 scale.
The last layer after the element-wise summation consists of $1 \times 1$ convolution, group norm and ReLU.
It should be noted that we
feed normalized pixel coordinates to the deepest FPN level (at
1/32 scale), before the convolutions and bilinear upsamplings. 
The provided accurate position information is important for enabling position sensitivity and predicting instance-aware features.
Compared with the separated alternative, the unified mask feature representation is more effective and time efficient.

\begin{figure}[tbp]
\centering
\includegraphics[width=0.807\linewidth]{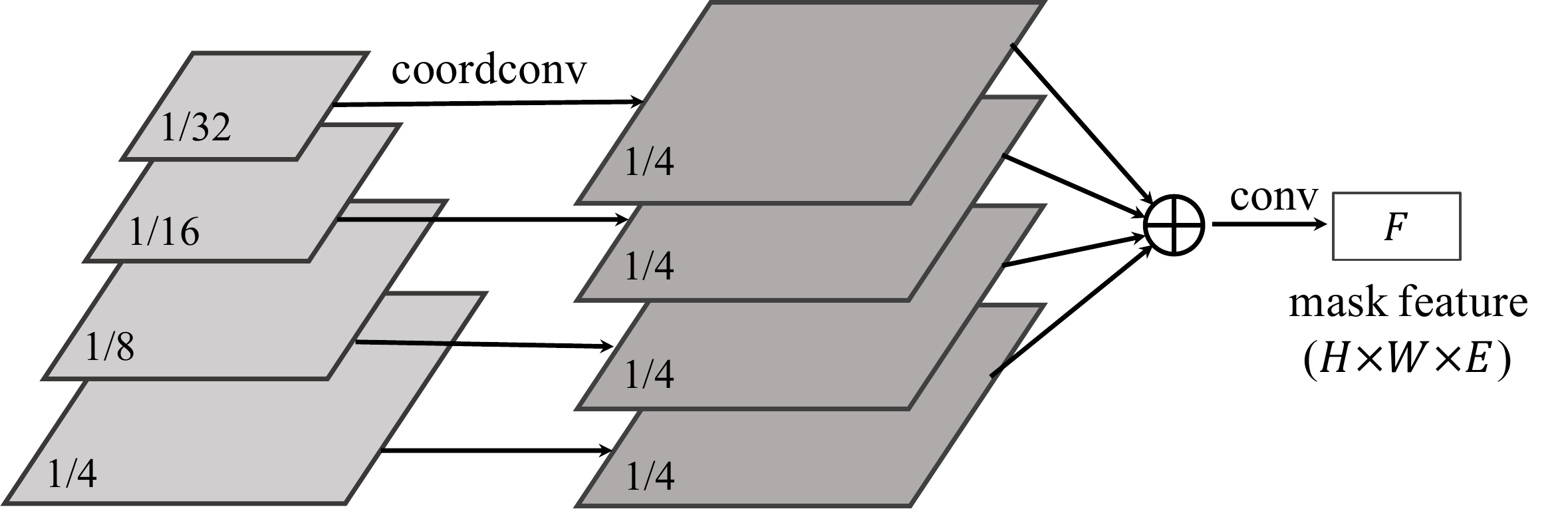}
\label{fig:mask_feature}
\caption{\textbf{Unified mask feature branch}. Each FPN level
(left) is upsampled by convolutions and bilinear upsampling
until it reaches 1/4 scale (middle).
In the deepest FPN level, we concatenate the $x$, $y$ coordinates and the original features to encode spatial information. 
After element-wise summation, a $1 \times 1$ convolution is attached to transform to designated output mask feature $F \in {\mathbb R}^{H\times W\times E}$. 
}
\label{fig:mask_feature}
\end{figure}

\section{Visualization}
We visualize what our \OurMethod has learnt
from two aspects: mask feature behavior and the final outputs after being convolved by the 
% predicted 
dynamically learned 
convolution kernels.

\begin{figure*}[b]
\begin{center}
    \includegraphics[width=0.98\linewidth]{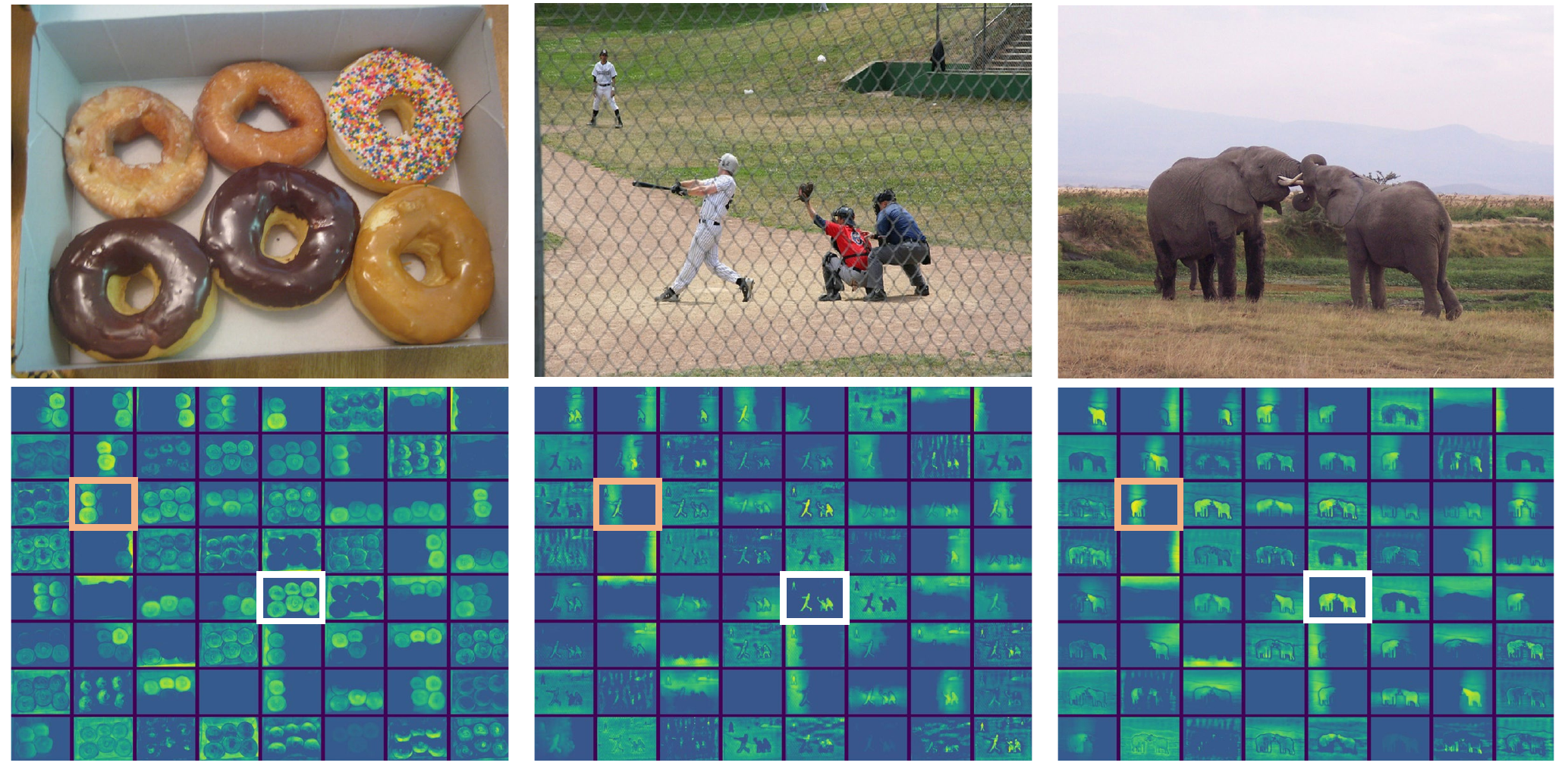}
\end{center}
   \caption{\textbf{Mask Feature Behavior.} 
   Each plotted subfigure corresponds to one of the 64 channels of the last feature map prior to mask prediction.
   %
   % Most of 
   The mask features 
   % are
   appear to be
   position-sensitive (orange box), while a few mask features are position-agnostic and activated on all instances (white box). Best viewed on screens.}
\label{fig:behavior}
\end{figure*}

% mask base visualization 
We 
%show
visualize
the outputs of mask feature branch.
% in Figure~\ref{fig:behavior}.
We use a model which has 64 output channels  (\ie,  $E=64$ for the last feature map prior to mask prediction) for 
% clear
easy
visualization.
Here  we plot each of the 64 channels (recall the channel spatial resolution is $ H \times W $) as shown in Figure~\ref{fig:behavior}. 

There are two main patterns.
The first and the foremost, the mask features are position-aware.
It shows obvious behavior of scanning the objects in the image horizontally and vertically.
%Interestingly, it is 
% naturally 
%indeed 
%in accordance to the target in the decoupled-head  SOLO: Segmenting objects by their independent horizontal and vertical location categories. 
The other obvious pattern is that some feature maps are responsible for activating all the foreground objects, \eg, the one in white boxes.

The final outputs are shown in Figure~\ref{fig:Vis}.
Different objects are in different colors.
Our method shows promising results in diverse scenes.
It %should benoted 
is worth pointing out
that the details at the boundaries are segmented 
%quite
well, especially for 
%the
large objects.
% We compare against Mask R-CNN on object details in Figure~\ref{fig:details_compare}.
% Our method shows great advantages.

We also provide three videos for better visualization of our instance segmentation results. These videos are generated from frame-by-frame inference, without any temporal processing. Though only trained on MS COCO, our model generalizes well across various scenes.

\section{Bounding-box Object  Detection}

\begin{figure}
\centering 
    \includegraphics[width=0.55\linewidth]{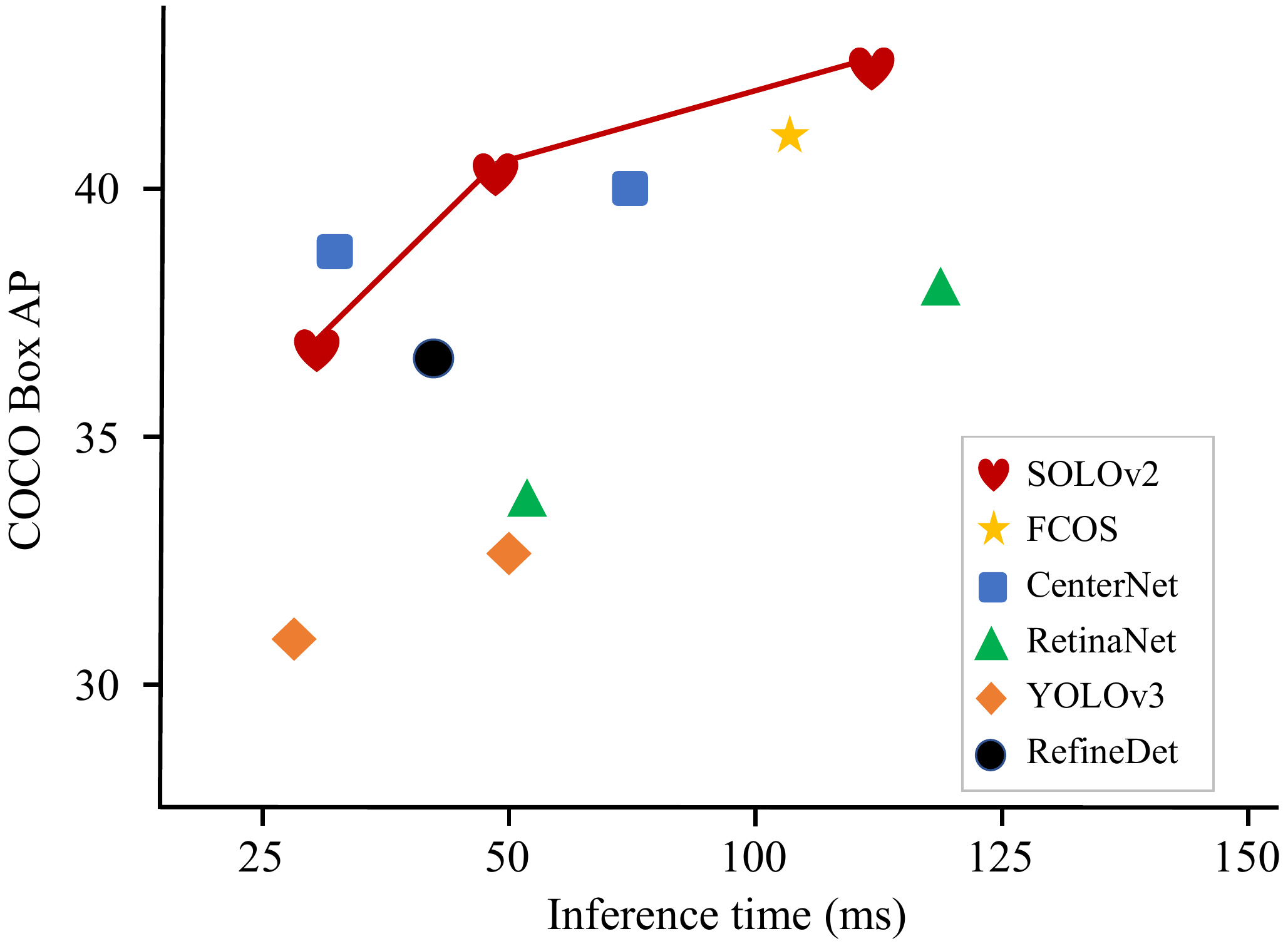}
\caption{Speed-accuracy trade-off of bounding-box \textbf{object detection} on the COCO \texttt{test}-\texttt{dev}.}
\label{fig:box_speed_ap}
\end{figure}

Although our instance segmentation solution removes the dependence of bounding box prediction, we are able to 
produce the 4D object bounding box from each instance mask. In Table~\ref{table:box_det}, we compare the generated box detection performance with other object detection methods on COCO. All models are trained on
the
{\tt train2017} subset and tested on 
{\tt test-dev.}

%We compare \Ours with other modern object detection methods on COCO. 
As shown in Table~\ref{table:box_det}, our detection results outperform most methods, especially for objects of large scales, demonstrating the effectiveness of \Ours in object box detection. Similar to instance segmentation, we also plot the speed/accuracy trade-off curve for different methods
in Figure~\ref{fig:box_speed_ap}.
We show our models with ResNet-101 and two light-weight versions described above.
The plot reveals that the bounding box performance of \Ours beats most recent object detection methods in both accuracy and speed. Here we emphasis that our results are directly generated from the off-the-shelf instance mask, without any box based supervised training or engineering.

An observation from Figure~\ref{fig:box_speed_ap} is as follows.
If one does not care much about the cost difference between mask  
annotation and bounding box annotation, it appears to us that there is no reason to use box detectors for downstream applications, considering the fact that our \Ours beats most modern detectors in both accuracy and speed.

\iffalse
\begin{figure}[t]
\begin{minipage}{0.646\linewidth}
\centering 
    \includegraphics[width=0.95\linewidth]{figures/box_speed_ap.pdf}
   \caption{\textbf{Speed-accuracy trade-off of 
   bounding-box
   object detection} on the COCO \texttt{test}-\texttt{dev}. 
   Inference time of all methods is 
   % tested
   evaluated 
   on a  TitanX-GPU machine.
   }
\label{fig:box_speed_ap}
\end{minipage}
\begin{minipage}{0.46\linewidth}
\begin{tabular}{ r |c|cc}
  &  PQ & PQ\textsuperscript{Th} & PQ\textsuperscript{St} \\
\Xhline{1pt}
\emph{box-based:} &&&\\
~~~~AUNet~\cite{li2019attention}       &  39.6 & 49.1 & 25.2  \\
~~~~UPSNet~\cite{xiong2019upsnet}        &  \textbf{42.5} & 48.5 & \textbf{33.4} \\

~~~~Panoptic-FPN~\cite{kirillov2019panoptic} &  39.0 & 45.9 & 28.7   \\
~~~~Panoptic-FPN$^*$-1\X      &  38.7 & 45.9 & 27.8  \\
~~~~Panoptic-FPN$^*$-3\X    & 40.8 & 48.3 & 29.4 \\
\hline\hline
\emph{box-free:} &&&\\
~~~~AdaptIS~\cite{adaptis}        & 35.9 & 40.3  & 29.3   \\
~~~~SSAP (Res101)~\cite{Gao_2019_ICCV}  & 36.5 & $-$ & $-$ \\
~~~~Pano-DeepLab (Xcept71) ~\cite{cheng2019panoptic}  & 39.7 & 43.9 & 33.2 \\
~~~~\textbf{\Ours}             & 42.1 & \textbf{49.6} & 30.7  \\
\end{tabular}
\end{minipage}
\end{figure}
\fi

\begin{table*}[ht]
\centering
\small
\begin{tabular}{ r  |l|cccccc}
  & backbone  & AP & AP$_{50}$ & AP$_{75}$ & AP$_S$ & AP$_M$ & AP$_L$  \\
\Xhline{1pt}
YOLOv3~\cite{yolov3}       & DarkNet53   & 33.0 & 57.9 & 34.4 & 18.3 & 35.4 & 41.9 \\
SSD513~\cite{ssd}          & ResNet-101  & 31.2 & 50.4 & 33.3 & 10.2 & 34.5 & 49.8 \\
DSSD513~\cite{ssd}         & ResNet-101  & 33.2 & 53.3 & 35.2 & 13.0 & 35.4 & 51.1 \\
RefineDet~\cite{refinedet} & ResNet-101  & 36.4 & 57.5 & 39.5 & 16.6 & 39.9 & 51.4 \\
Faster R-CNN~\cite{fpn}    & Res-101-FPN & 36.2 & 59.1 & 39.0 & 18.2 & 39.0 & 48.2 \\
RetinaNet~\cite{focalloss} & Res-101-FPN & 39.1 & 59.1 & 42.3 & 21.8 & 42.7 & 50.2 \\
FoveaBox~\cite{foveabox}      & Res-101-FPN & 40.6 & 60.1 & 43.5 & 23.3 & 45.2 & 54.5 \\
RPDet~\cite{reppoints}         & Res-101-FPN & 41.0 & 62.9 & 44.3 & 23.6 & 44.1 & 51.7 \\
FCOS~\cite{fcos}           & Res-101-FPN & 41.5 & 60.7 & 45.0 & \textbf{24.4} & 44.8 & 51.6 \\
CenterNet~\cite{centernet} &Hourglass-104& 42.1 & 61.1 & 45.9 & 24.1 & 45.5 & 52.8 \\
\hline
\textbf{\Ours}  & Res-50-FPN       & 40.4 & 59.8 & 42.8 & 20.5 & 44.2 & 53.9 \\
\textbf{\Ours}  & Res-101-FPN      & 42.6 & 61.2 & 45.6 & 22.3 & 46.7 & 56.3 \\
\textbf{\Ours}  & Res-DCN-101-FPN  & \textbf{44.9} & \textbf{63.8} & \textbf{48.2} & 23.1 & \textbf{48.9} & \textbf{61.2} \\
\end{tabular}
\caption{\textbf{Object detection} box AP (\%) on 
the
COCO \texttt{test}-\texttt{dev}. Although our bounding boxes are directly generated from the predicted masks, the accuracy outperforms most state-of-the-art methods. Speed-accuracy trade-off of typical methods 
%are
is 
shown in Figure~\ref{fig:box_speed_ap}.
%
% ADD FCOS's result
% xl: fcos doesn't have 3x results on val set.
}
\label{table:box_det}
\end{table*}

\section{Results on the LVIS dataset}
LVIS~\cite{lvis2019} is 
a
recently proposed dataset for long-tail object segmentation, which has more than 1000 object categories. In LVIS, each object instance is segmented with a
high-quality mask that surpasses the annotation quality of
the relevant COCO dataset. Since LVIS is new, only the results of Mask R-CNN are publicly available. Therefore we only compare \Ours against the Mask R-CNN baseline. 

Table~\ref{table:lvis} 
reports the performances on the 
rare (1$\sim$10 images), common (11$\sim$100), and frequent ($>$ 100) subsets, as well as the overall AP. 
Both the reported Mask R-CNN and \Ours use data resampling training strategy, following~\cite{lvis2019}.
Our \Ours outperforms the baseline method by about 1\% AP. For large-size  objects (AP$_L$), our \Ours 
achieves 
6.7\%  AP 
improvement, which is consistent with the results on the COCO dataset.

\begin{table*}[t]
\centering
\small
\begin{tabular}{ r  |l|ccc|ccc|c}
  & backbone & AP$_r$ & AP$_c$ & AP$_f$ & AP$_S$ & AP$_M$ & AP$_L$&AP \\
\Xhline{1pt}
Mask-RCNN~\cite{lvis2019}& Res-50-FPN     & 14.5 & 24.3 & 28.4 & - & - & - & 24.4 \\
Mask-RCNN$^*$-3\X & Res-50-FPN               & 12.1 & 25.8 & 28.1 & \textbf{18.7} & 31.2 & 38.2 & 24.6 \\
\hline
\textbf{\Ours}  & Res-50-FPN      & 13.4 & 26.6 & 28.9 & 15.9 & 34.6 & 44.9 & 25.5 \\
\textbf{\Ours}  & Res-101-FPN     & \textbf{16.3} & \textbf{27.6} & \textbf{30.1} & 16.8 & \textbf{35.8} & \textbf{47.0} & \textbf{26.8}
\end{tabular}
\caption{Instance segmentation results on the 
{ LVIS}v0.5 validation dataset.  
}
\label{table:lvis}
\end{table*}

\section{Panoptic Segmentation}
% TODO: modify
The proposed \Ours can be easily extended to panoptic segmentation by adding the semantic segmentation branch, analogue to the mask feature branch.
%We use the semantic segmentation branch of PanopticFPN to extend \Ours to the panoptic segmentation task. 
We use annotations of COCO 2018 panoptic segmentaiton task. All models are trained on {\tt train2017}
subset and tested on {\tt  val2017}. 
We use the same strategy as in Panoptic-FPN to combine instance and semantic results. 
As shown in Table~\ref{table:panoptic}, our method achieves state-of-the-art results and outperforms  other recent box-free methods by a large margin. All methods listed use the same backbone (ResNet50-FPN) except SSAP (ResNet101) and Panoptic-DeepLab (Xception-71).

\begin{table}[ht]
\centering
\small
\begin{tabular}{ r |c|cc}
  &  PQ & PQ\textsuperscript{Th} & PQ\textsuperscript{St} \\
\Xhline{1pt}
\emph{box-based:} &&&\\
~~~~AUNet~\cite{li2019attention}       &  39.6 & 49.1 & 25.2  \\
~~~~UPSNet~\cite{xiong2019upsnet}        &  \textbf{42.5} & 48.5 & \textbf{33.4} \\

~~~~Panoptic-FPN~\cite{kirillov2019panoptic} &  39.0 & 45.9 & 28.7   \\
~~~~Panoptic-FPN$^*$-1\X      &  38.7 & 45.9 & 27.8  \\
~~~~Panoptic-FPN$^*$-3\X    & 40.8 & 48.3 & 29.4 \\
\hline\hline
\emph{box-free:} &&&\\
~~~~AdaptIS~\cite{adaptis}        & 35.9 & 40.3  & 29.3   \\
~~~~SSAP (Res101)~\cite{Gao_2019_ICCV}  & 36.5 & $-$ & $-$ \\
~~~~Pano-DeepLab (Xcept71) ~\cite{cheng2019panoptic}  & 39.7 & 43.9 & 33.2 \\
~~~~\textbf{\Ours}             & 42.1 & \textbf{49.6} & 30.7  \\
\end{tabular}
\caption{Panoptic results on COCO val2017. Here Panoptic-FPN$^*$ is our re-implemented version in \texttt{mmdetection}~\cite{mmdetection} with 12 and 36 training epochs (1\X \, and 3\X) and multi-scale training. All model's backbones are Res50, except SSAP and Pano-DeepLab.
Note that 
UPSNet has used deformable convolution~\cite{dai2017deformable} for better performance.}
\label{table:panoptic}
\end{table}

\def\visimgheighta{2.84cm}
\def\visimgwidtha{3.86cm}

\def\visimgheightb{2.74cm}
\def\visimgwidtb{3.86cm}

\def\visimgheightc{3.04cm}
\def\visimgwidtc{3.86cm}

\def\visimgheightd{3.16cm}
\def\visimgwidtd{3.86cm}

\def\visimgheighte{2.68cm} 
\def\visimgwidthe{3.86cm}

\def\visimgheightf{3.15cm}
\def\visimgwidthf{3.86cm}

\def\visimgheightg{3.34cm}
\def\visimgwidthg{3.86cm}

\begin{figure*}[ht]
\centering 
{
\includegraphics[height=\visimgheighta]{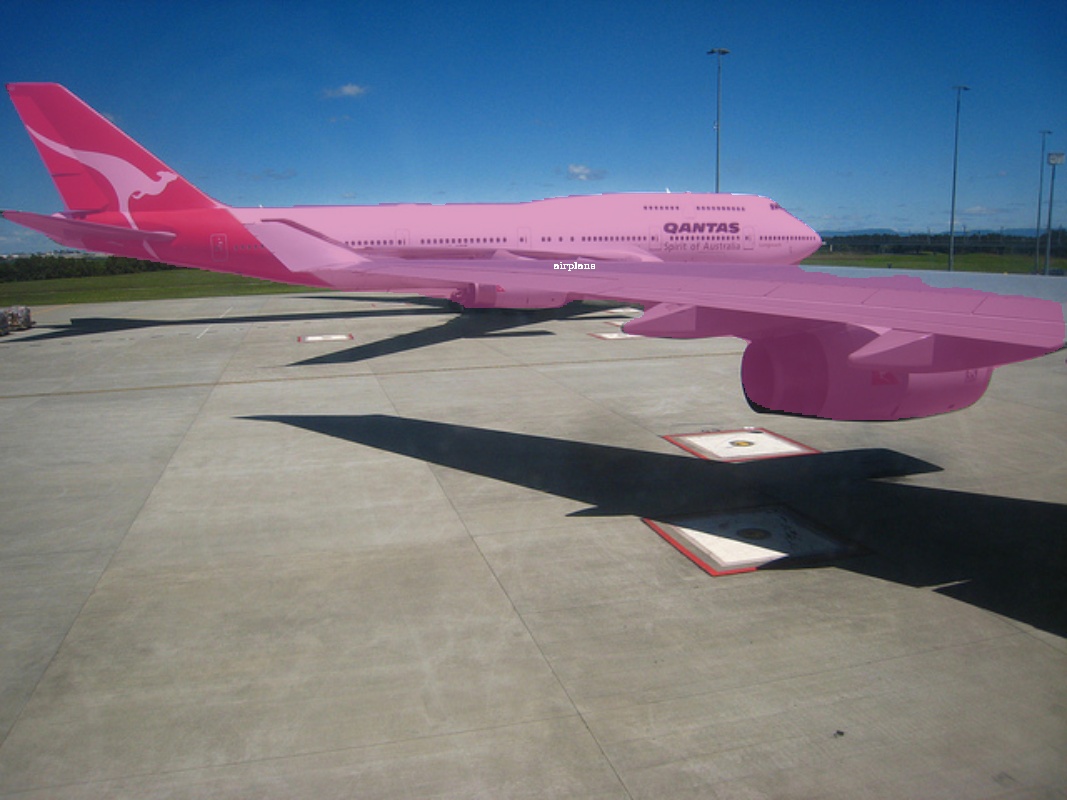}
\includegraphics[height=\visimgheighta]{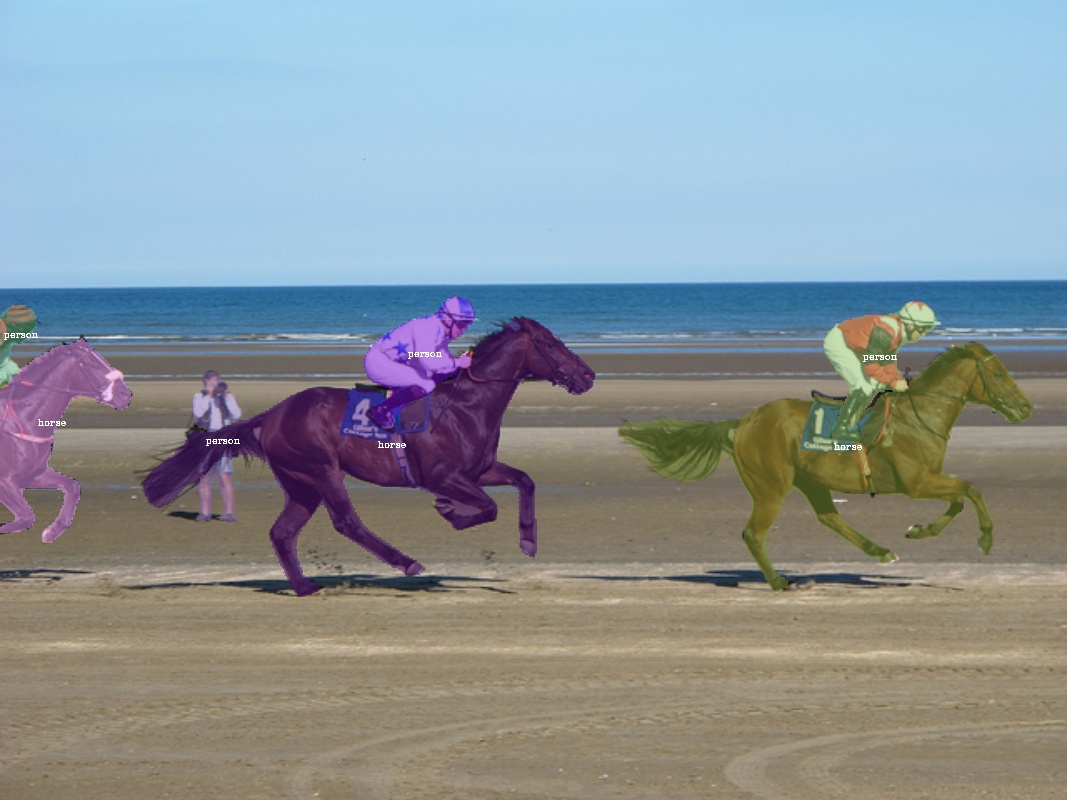}
\includegraphics[height=\visimgheighta]{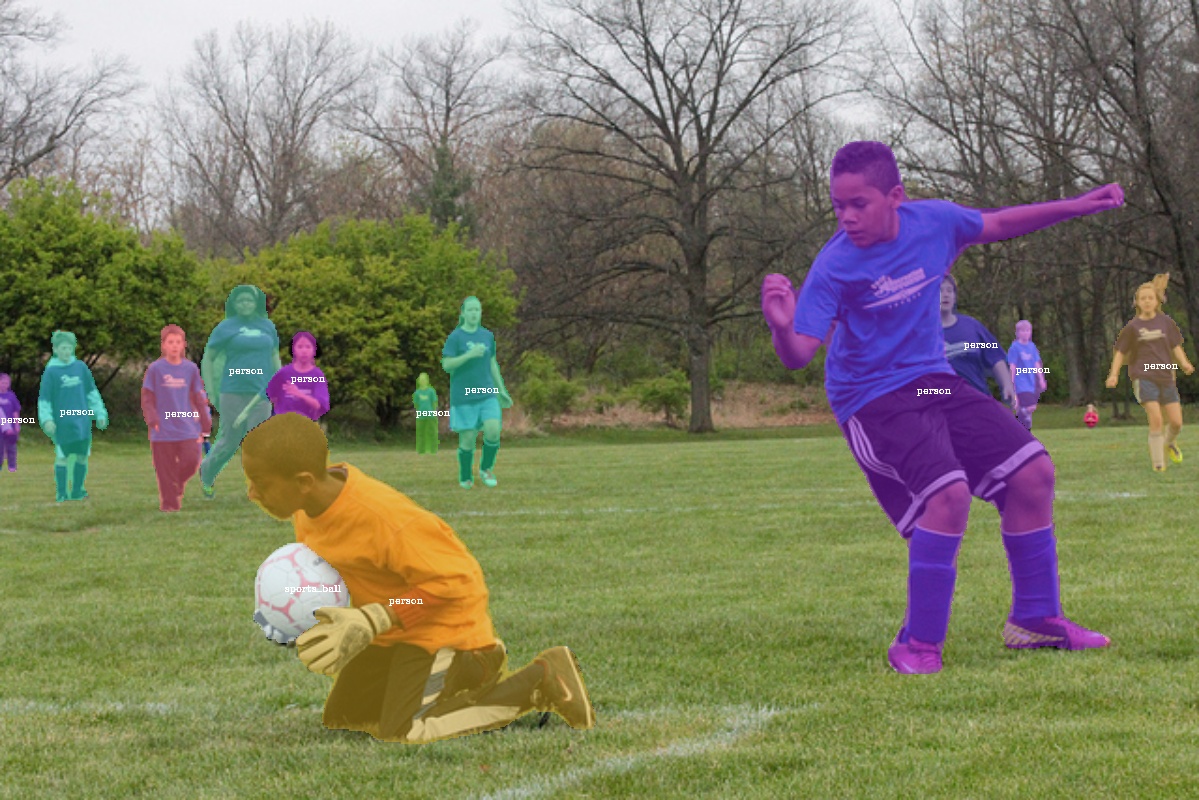}
\includegraphics[height=\visimgheighta]{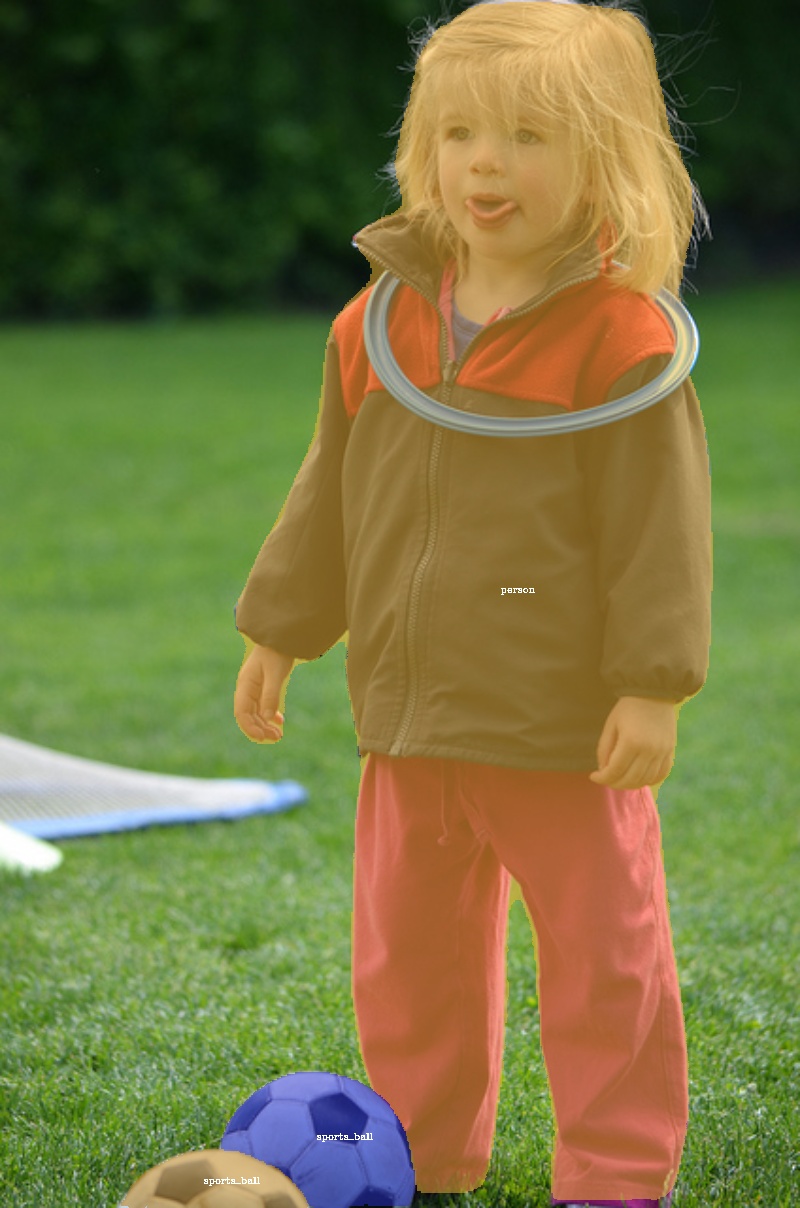}

\includegraphics[height=\visimgheightb]{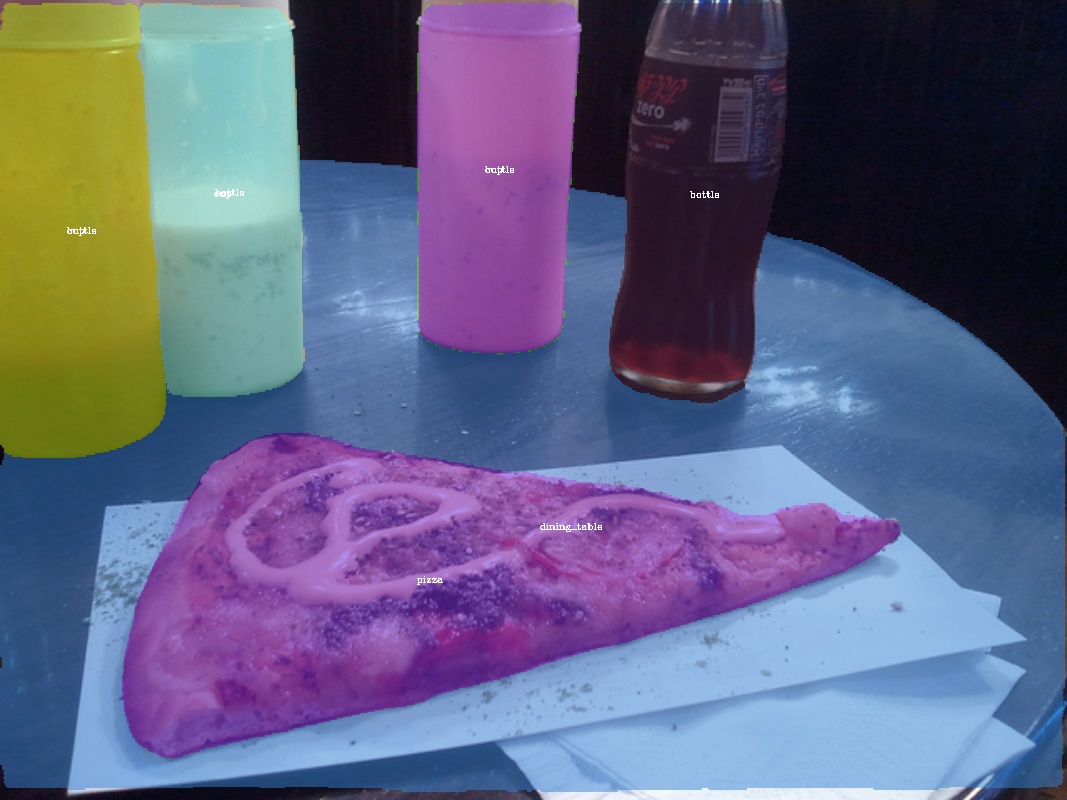}
\includegraphics[height=\visimgheightb]{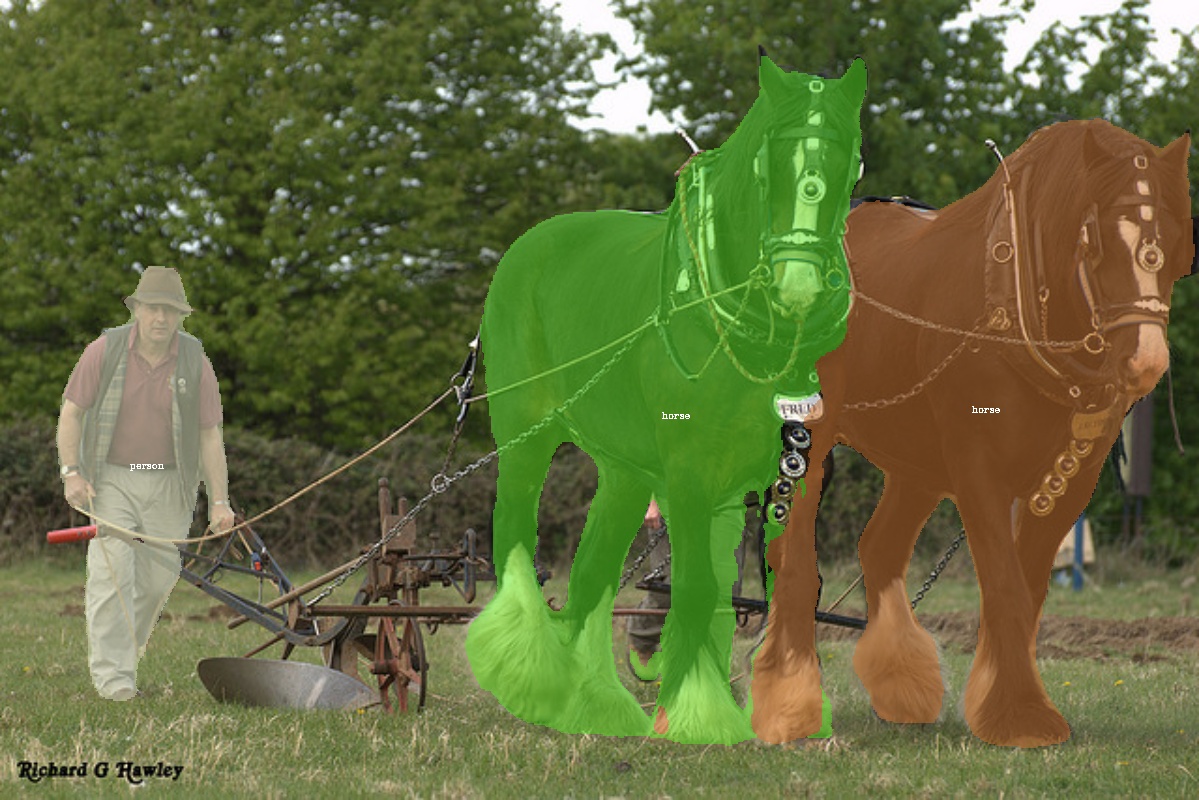}
\includegraphics[height=\visimgheightb]{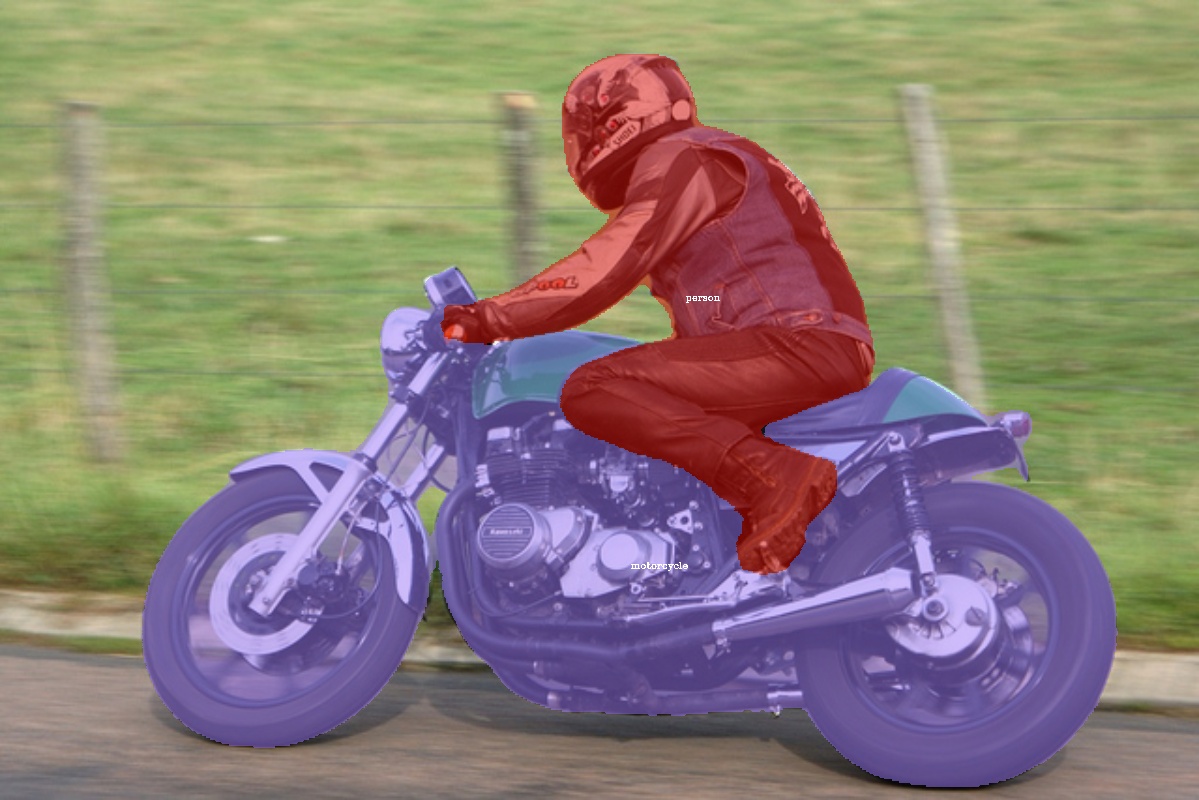}
\includegraphics[height=\visimgheightb]{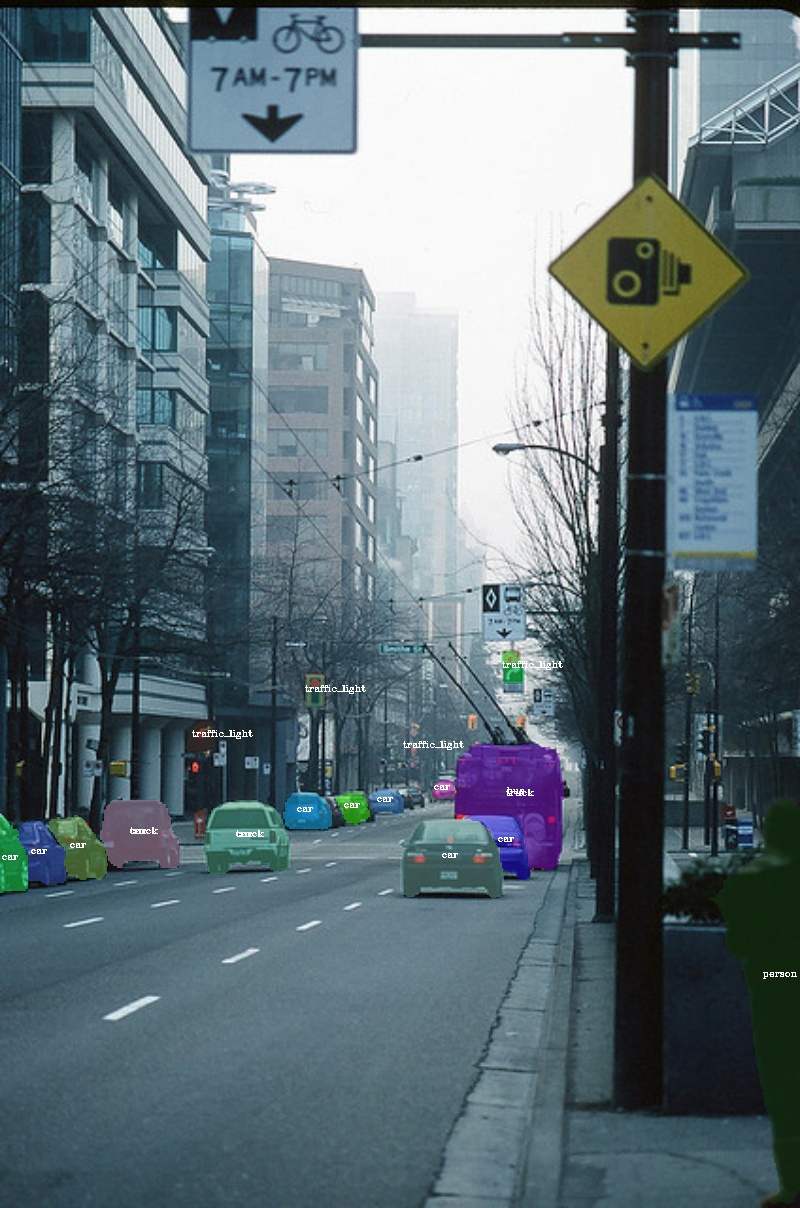}

\includegraphics[height=\visimgheightc]{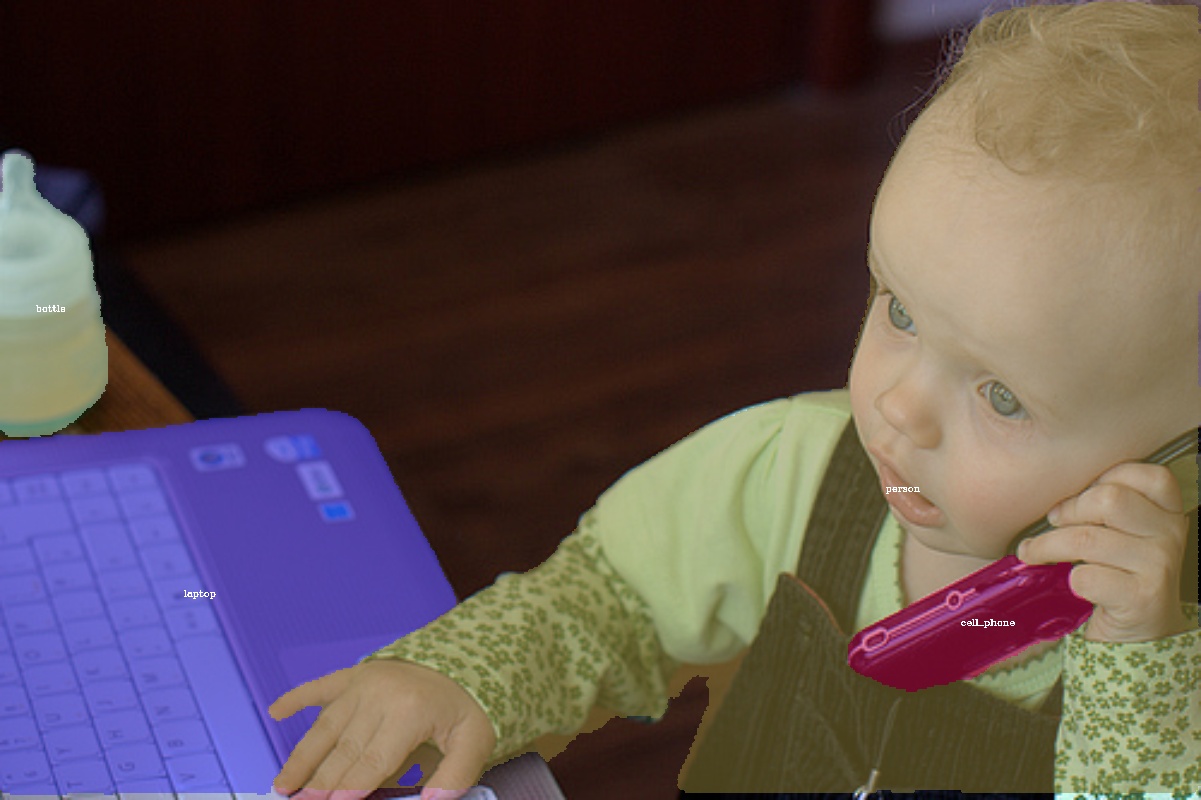}
\includegraphics[height=\visimgheightc]{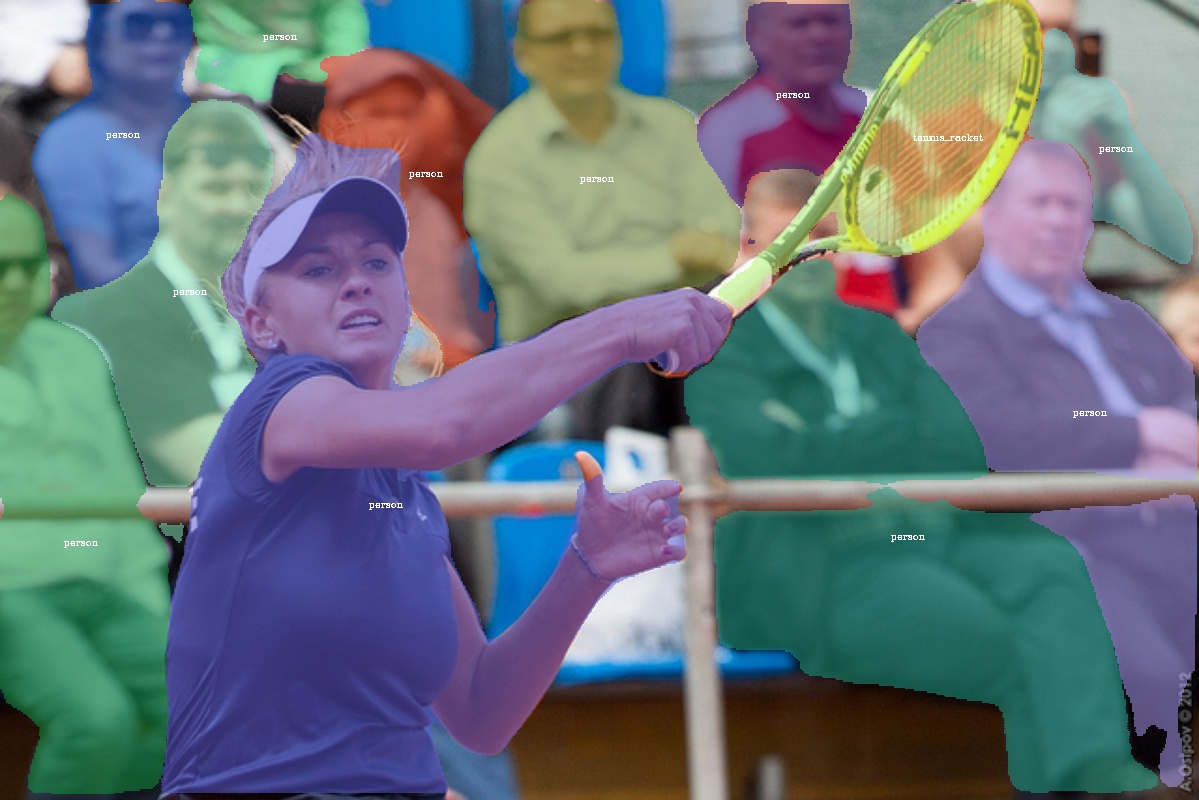}
\includegraphics[height=\visimgheightc]{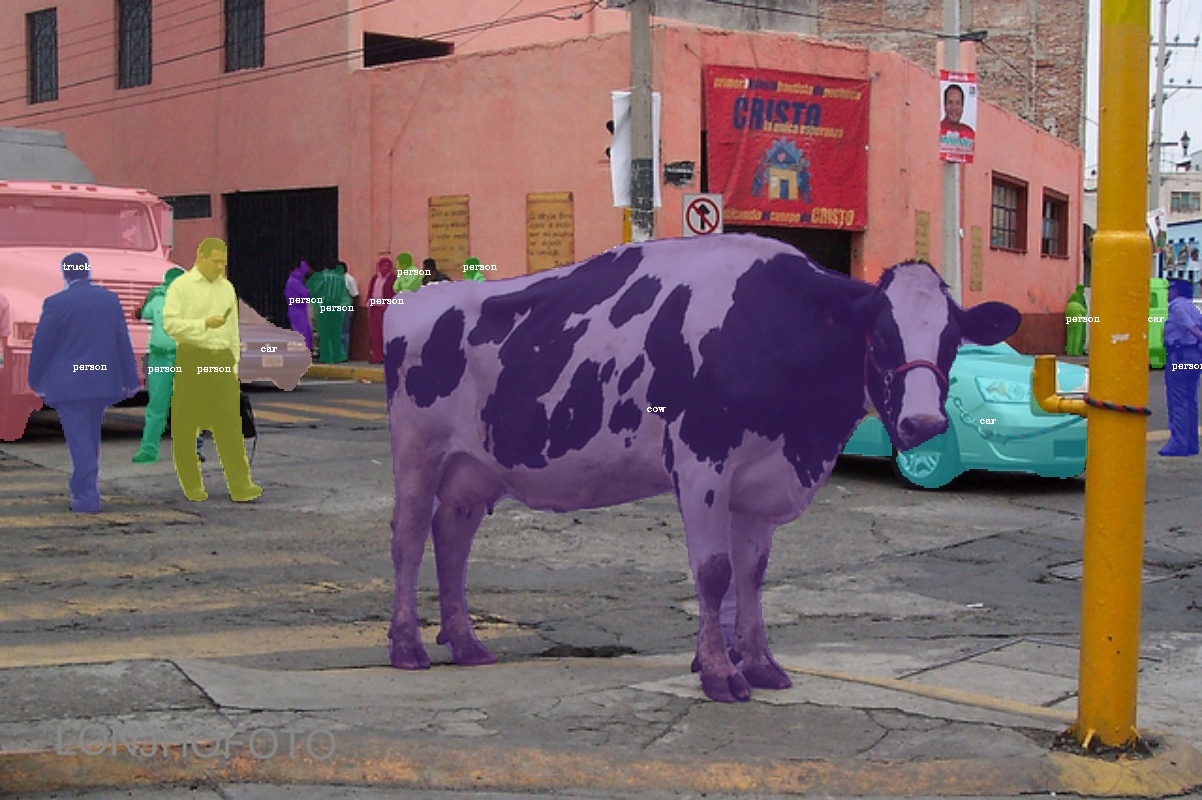}

\includegraphics[height=\visimgheightd]{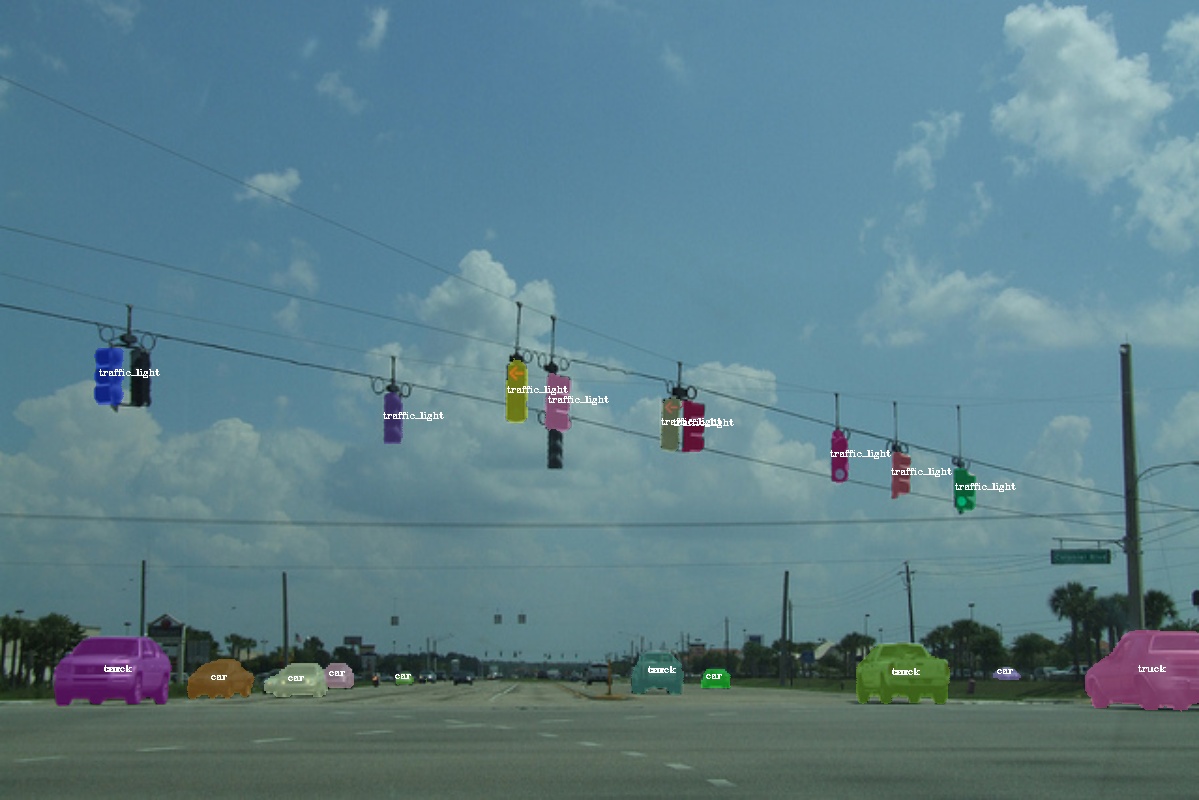}
\includegraphics[height=\visimgheightd]{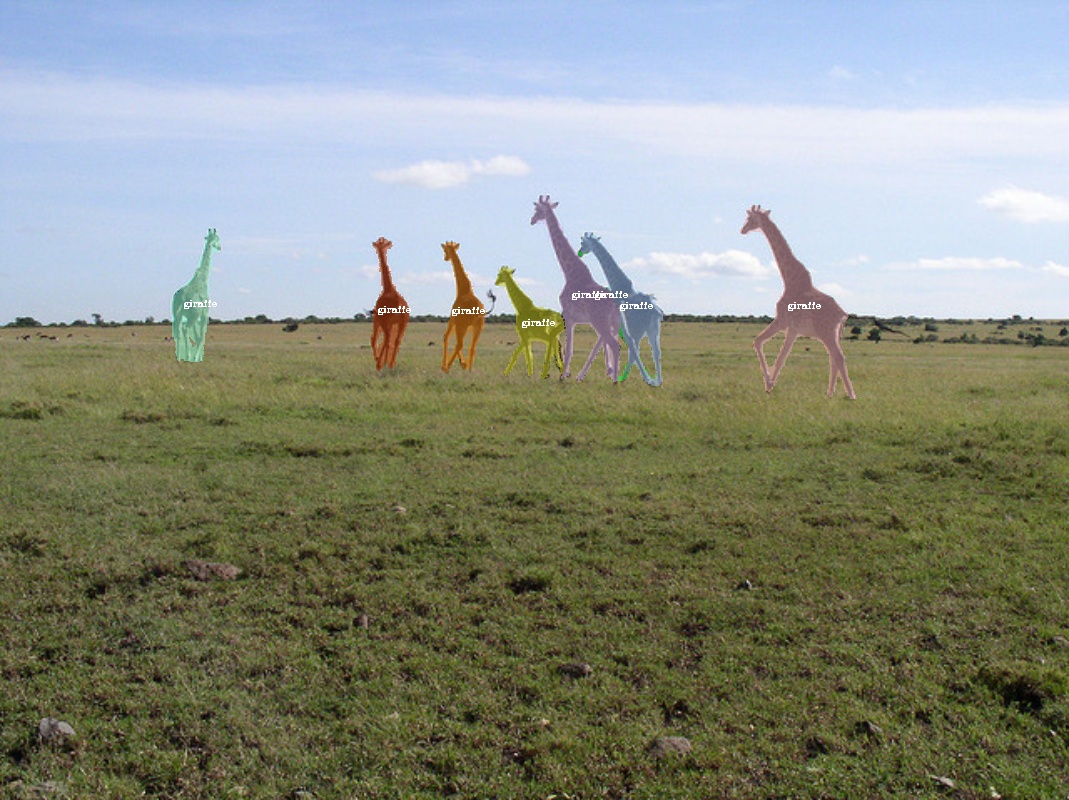}
\includegraphics[height=\visimgheightd]{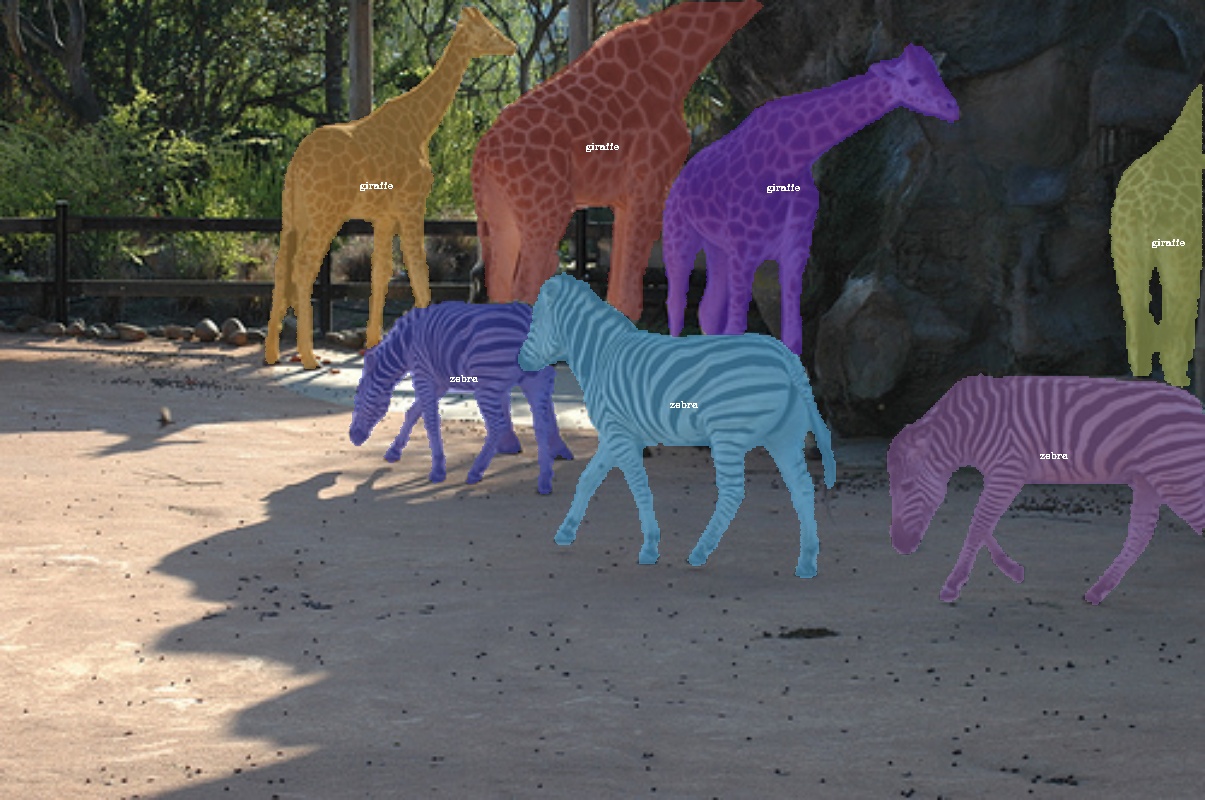}

\includegraphics[height=\visimgheighte]{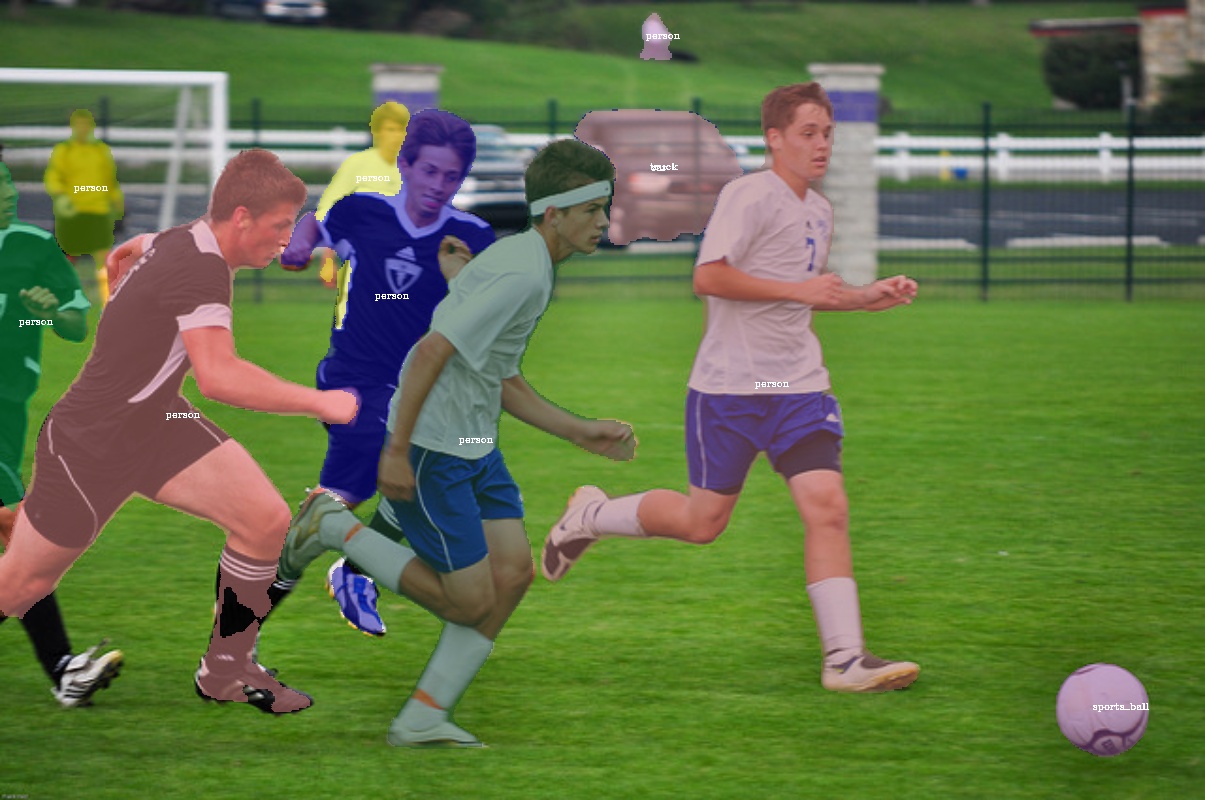}
\includegraphics[height=\visimgheighte]{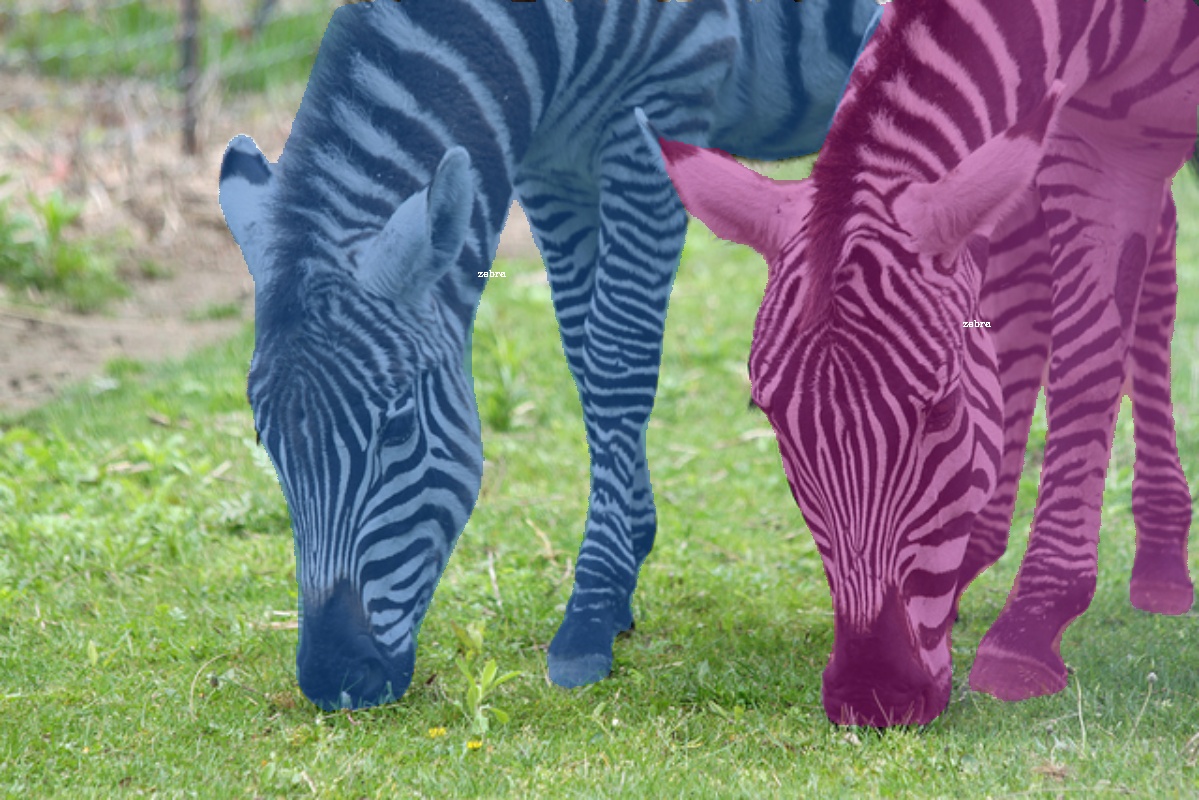}
\includegraphics[height=\visimgheighte]{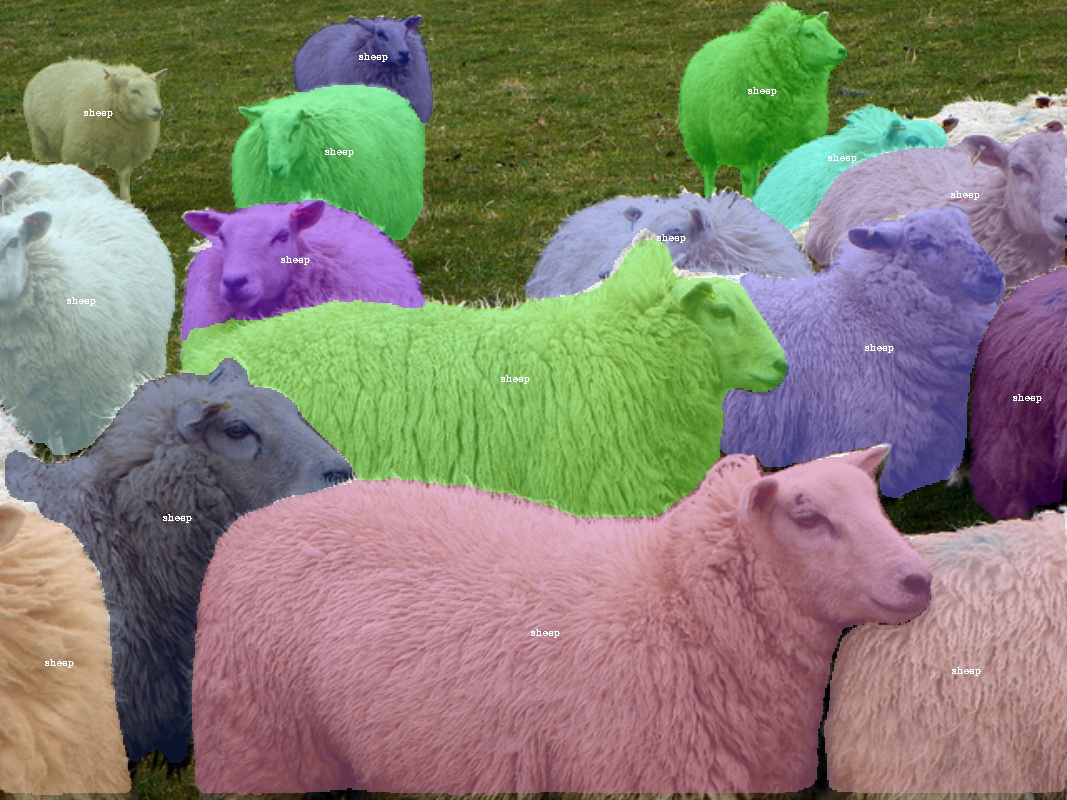}
\includegraphics[height=\visimgheighte]{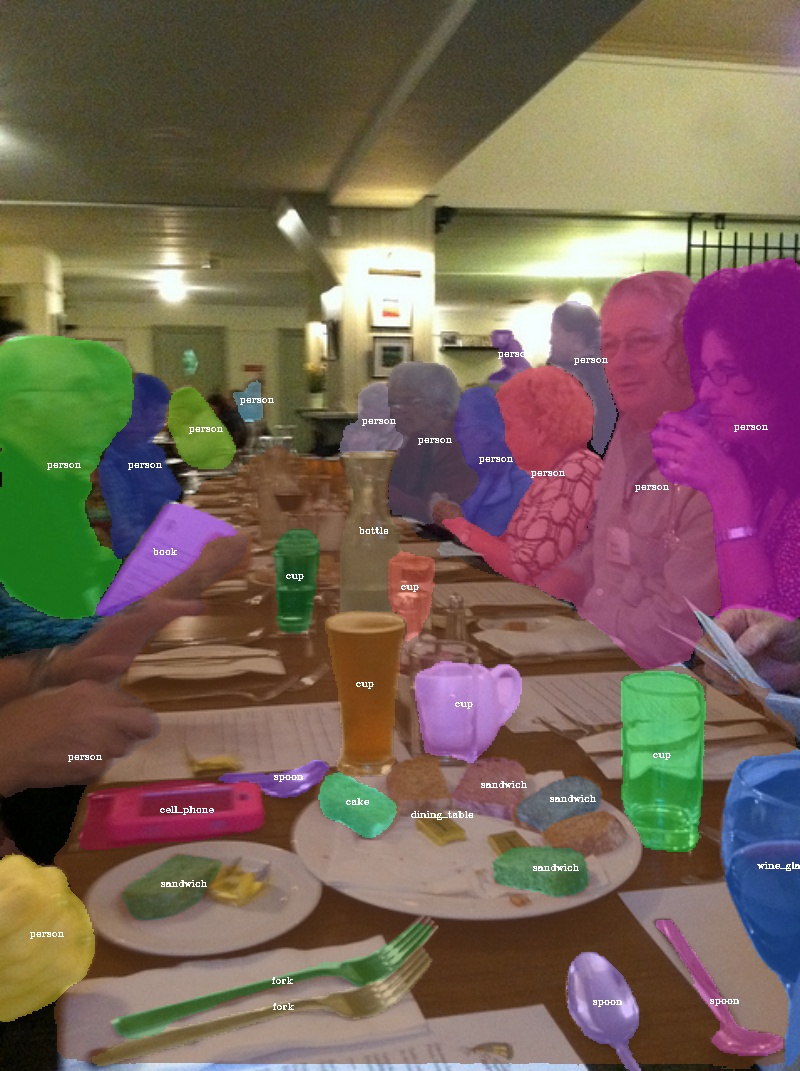}

\includegraphics[height=\visimgheightf]{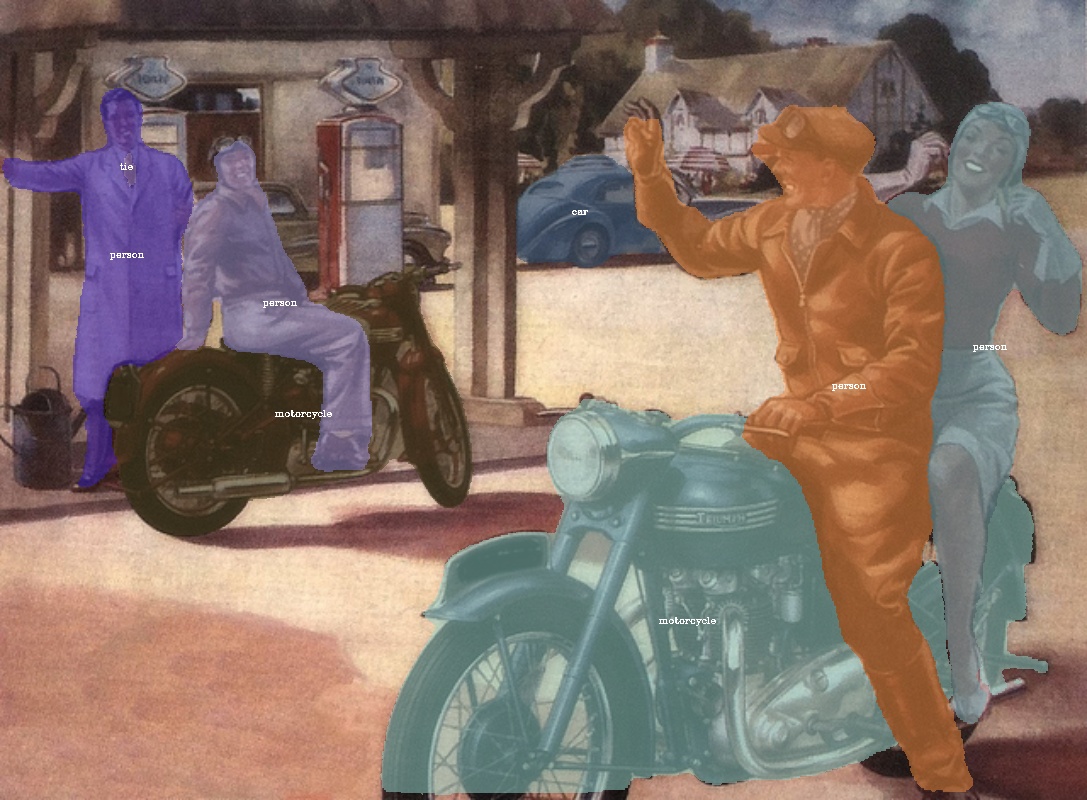}
\includegraphics[height=\visimgheightf]{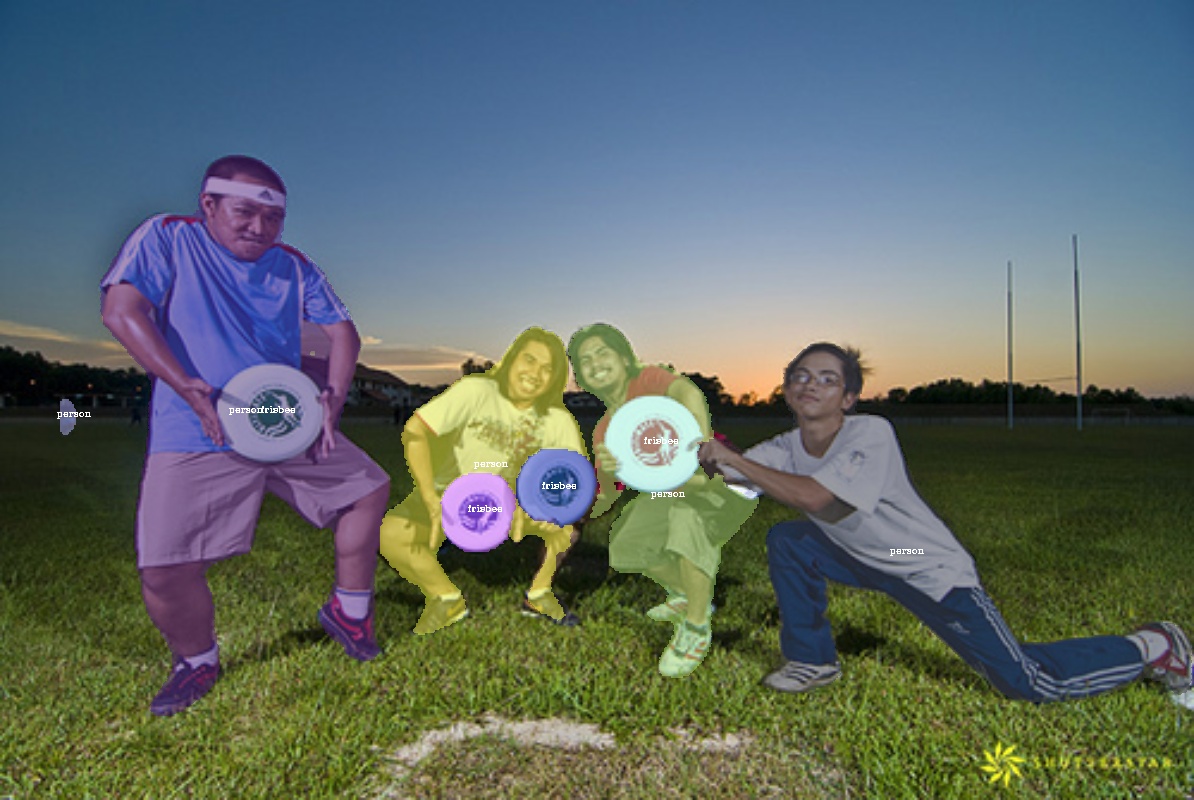}
\includegraphics[height=\visimgheightf]{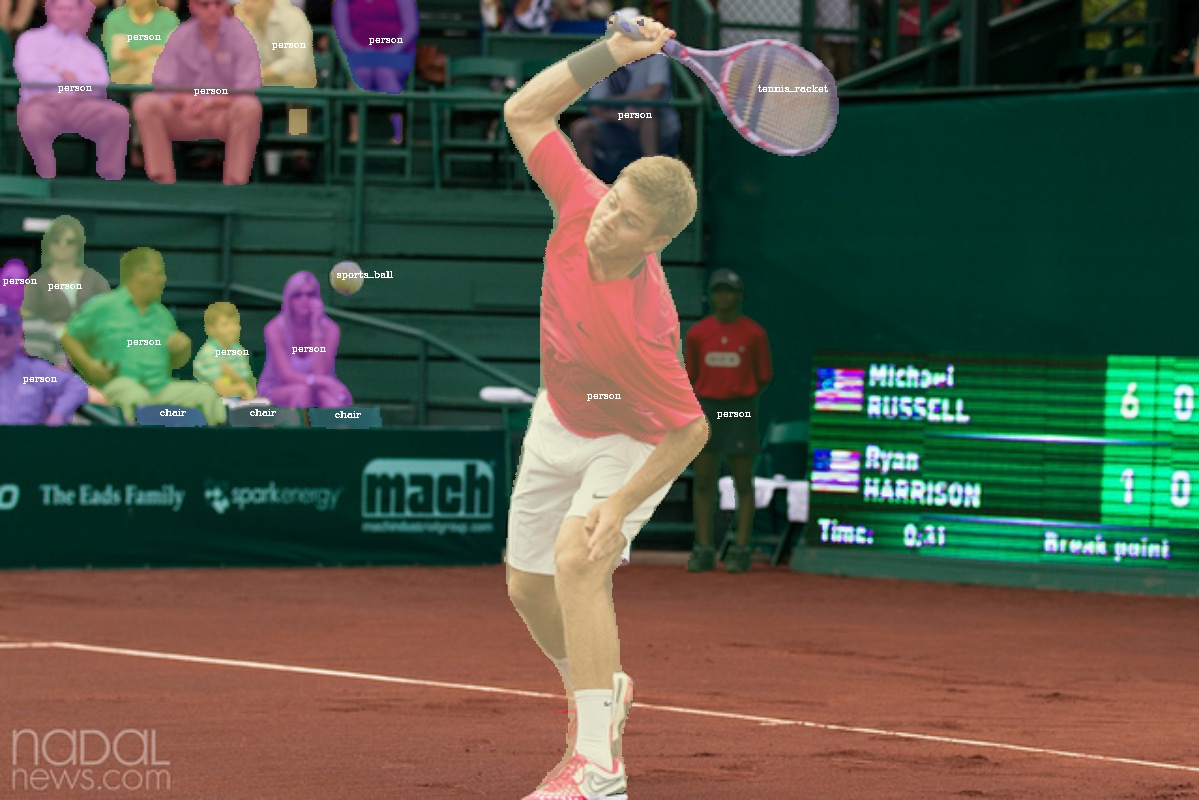}

\includegraphics[height=\visimgheightg]{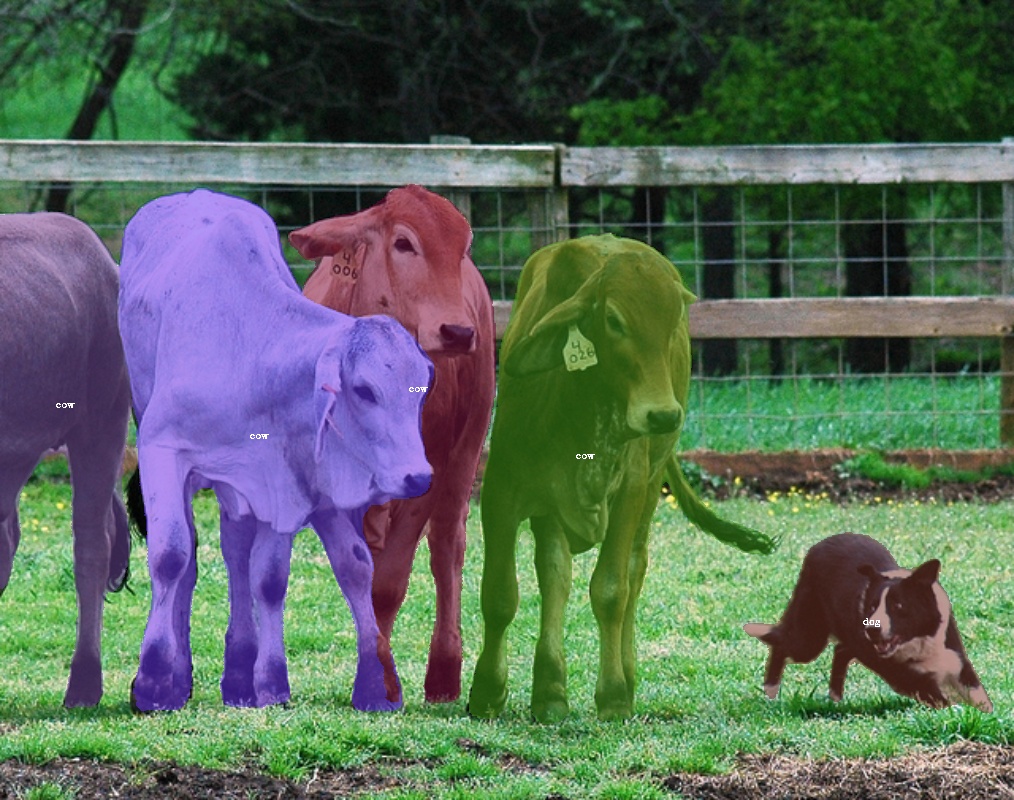}
\includegraphics[height=\visimgheightg]{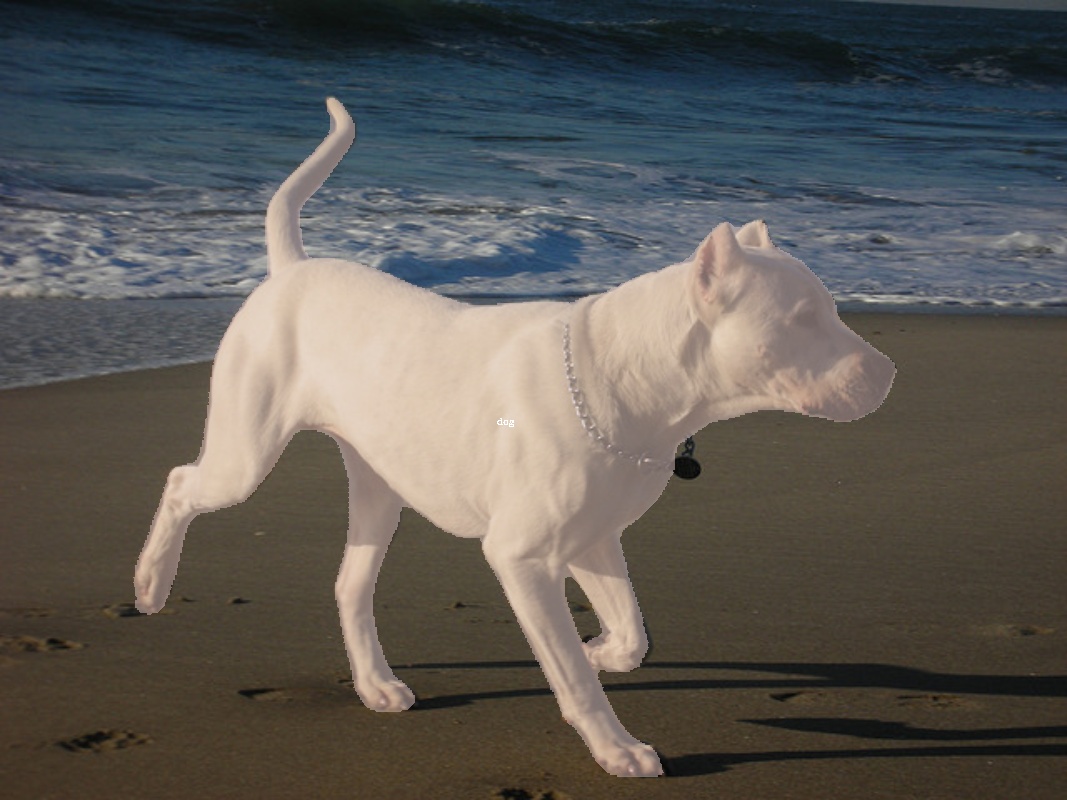}
\includegraphics[height=\visimgheightg]{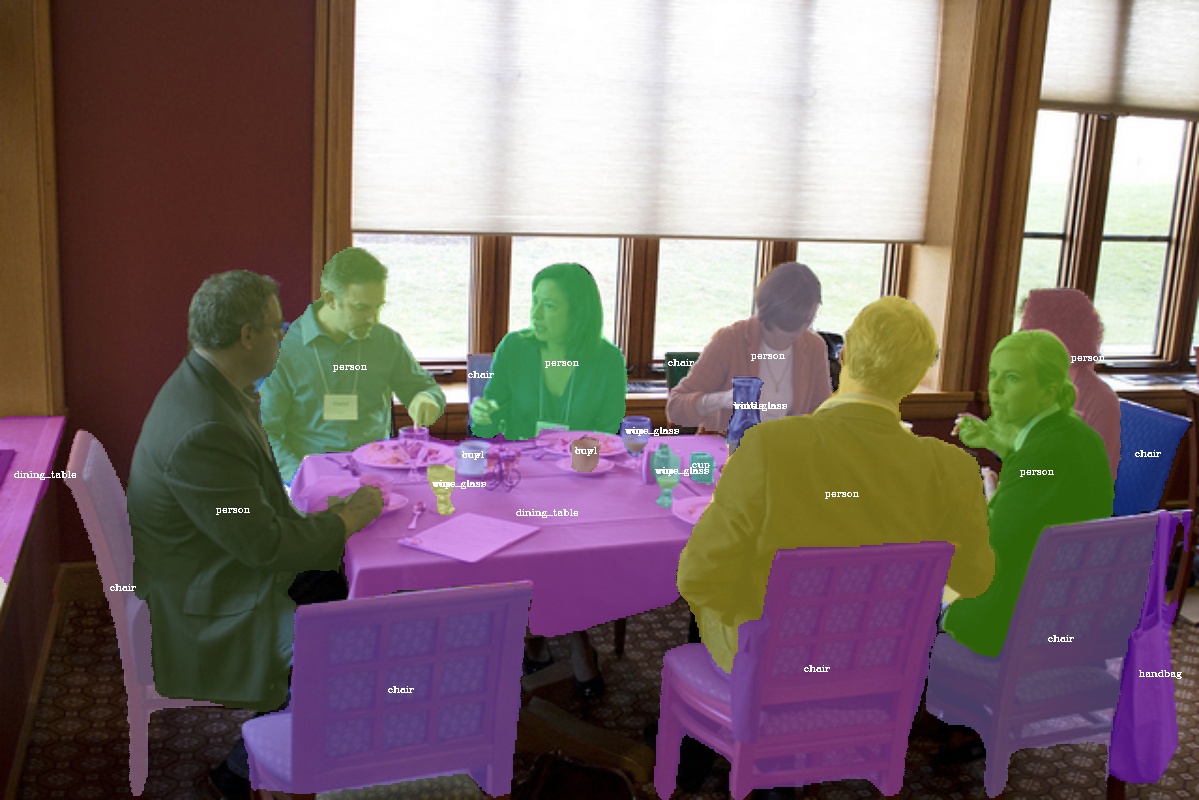}

} % scalebox
\caption{\textbf{Visualization of instance segmentation results}
using the Res-101-FPN backbone.
The model is trained on the COCO \texttt{train2017} dataset, achieving a mask AP of 39.7\%  on the COCO \texttt{test}-\texttt{dev}.
}
\label{fig:Vis}
\end{figure*}

% details comparison

\clearpage

\section*{Broader Impact}

One of the primary goals of computer vision is %
understanding of visual scenes. Scene understanding involves numerous tasks (\eg, recognition, detection, segmentation, \textit{etc}.). Among them, instance segmentation is probably one of the most challenging tasks, which requires to 
detect object instances at the pixel level.

Albeit 
being challenging, instance segmentation is beneficial %
to a wide range of applications, including autonomous driving, augmented reality, medical image analysis, and %
image/video editing.
The proposed accurate and fast instance segmentation solution benefits broader applications. Autonomous driving becomes %
safer 
Doctors could find the lesion part in medical images with less effort. 

Moreover, we believe that
our method can serve as a strong baseline for researchers and engineers in the field.
This new paradigm may encourage future work to deeply analyze and further enhance research along this direction. 
Practitioners may develop interesting applications built upon our approach.

{\small
\bibliographystyle{unsrt}
\bibliography{solov2}

\begin{thebibliography}{10}

\bibitem{solo}
Xinlong Wang, Tao Kong, Chunhua Shen, Yuning Jiang, and Lei Li.
\newblock {SOLO}: Segmenting objects by locations.
\newblock In {\em Proc. Eur. Conf. Comp. Vis.}, 2020.

\bibitem{yolact}
Daniel Bolya, Chong Zhou, Fanyi Xiao, and Yong~Jae Lee.
\newblock {YOLACT}: {Real-time} instance segmentation.
\newblock In {\em Proc. IEEE Int. Conf. Comp. Vis.}, 2019.

\bibitem{fcis}
Yi~Li, Haozhi Qi, Jifeng Dai, Xiangyang Ji, and Yichen Wei.
\newblock Fully convolutional instance-aware semantic segmentation.
\newblock In {\em Proc. IEEE Conf. Comp. Vis. Patt. Recogn.}, 2017.

\bibitem{maskrcnn}
Kaiming He, Georgia Gkioxari, Piotr Doll{\'{a}}r, and Ross~B. Girshick.
\newblock Mask {R-CNN}.
\newblock In {\em Proc. IEEE Int. Conf. Comp. Vis.}, 2017.

\bibitem{panet}
Shu Liu, Lu~Qi, Haifang Qin, Jianping Shi, and Jiaya Jia.
\newblock Path aggregation network for instance segmentation.
\newblock In {\em Proc. IEEE Conf. Comp. Vis. Patt. Recogn.}, 2018.

\bibitem{maskscoringrcnn}
Zhaojin Huang, Lichao Huang, Yongchao Gong, Chang Huang, and Xinggang Wang.
\newblock Mask scoring {R-CNN}.
\newblock In {\em Proc. IEEE Conf. Comp. Vis. Patt. Recogn.}, 2019.

\bibitem{Chen_2019_ICCV}
Xinlei Chen, Ross Girshick, Kaiming He, and Piotr Dollar.
\newblock {TensorMask}: A foundation for dense object segmentation.
\newblock In {\em Proc. IEEE Int. Conf. Comp. Vis.}, 2019.

\bibitem{chen2020blendmask}
Hao Chen, Kunyang Sun, Zhi Tian, Chunhua Shen, Yongming Huang, and Youliang
  Yan.
\newblock {BlendMask}: Top-down meets bottom-up for instance segmentation.
\newblock In {\em Proc. IEEE Conf. Comp. Vis. Patt. Recogn.}, 2020.

\bibitem{MEInst}
Rufeng Zhang, Zhi Tian, Chunhua Shen, Mingyu You, and Youliang Yan.
\newblock Mask encoding for single shot instance segmentation.
\newblock In {\em Proc. IEEE Conf. Comp. Vis. Patt. Recogn.}, 2020.

\bibitem{polarmask}
Enze Xie, Peize Sun, Xiaoge Song, Wenhai Wang, Xuebo Liu, Ding Liang, Chunhua
  Shen, and Ping Luo.
\newblock {PolarMask}: Single shot instance segmentation with polar
  representation.
\newblock In {\em Proc. IEEE Conf. Comp. Vis. Patt. Recogn.}, 2020.

\bibitem{fcos}
Zhi Tian, Chunhua Shen, Hao Chen, and Tong He.
\newblock {FCOS}: Fully convolutional one-stage object detection.
\newblock In {\em Proc. IEEE Int. Conf. Comp. Vis.}, 2019.

\bibitem{associativeembedding}
Alejandro Newell, Zhiao Huang, and Jia Deng.
\newblock Associative embedding: End-to-end learning for joint detection and
  grouping.
\newblock In {\em Proc. Advances in Neural Inf. Process. Syst.}, 2017.

\bibitem{de2017semantic}
Bert De~Brabandere, Davy Neven, and Luc Van~Gool.
\newblock Semantic instance segmentation with a discriminative loss function.
\newblock {\em arXiv:1708.02551}, 2017.

\bibitem{SGN17}
Shu Liu, Jiaya Jia, Sanja Fidler, and Raquel Urtasun.
\newblock Sequential grouping networks for instance segmentation.
\newblock In {\em Proc. IEEE Int. Conf. Comp. Vis.}, 2017.

\bibitem{Gao_2019_ICCV}
Naiyu Gao, Yanhu Shan, Yupei Wang, Xin Zhao, Yinan Yu, Ming Yang, and Kaiqi
  Huang.
\newblock {SSAP}: Single-shot instance segmentation with affinity pyramid.
\newblock In {\em Proc. IEEE Int. Conf. Comp. Vis.}, 2019.

\bibitem{jaderberg2015spatial}
Max Jaderberg, Karen Simonyan, Andrew Zisserman, and Koray Kavukcuoglu.
\newblock Spatial transformer networks.
\newblock In {\em Proc. Advances in Neural Inf. Process. Syst.}, 2015.

\bibitem{jia2016dynamic}
Xu~Jia, Bert De~Brabandere, Tinne Tuytelaars, and Luc~Van Gool.
\newblock Dynamic filter networks.
\newblock In {\em Proc. Advances in Neural Inf. Process. Syst.}, 2016.

\bibitem{dai2017deformable}
Jifeng Dai, Haozhi Qi, Yuwen Xiong, Yi~Li, Guodong Zhang, Han Hu, and Yichen
  Wei.
\newblock Deformable convolutional networks.
\newblock In {\em Proc. IEEE Int. Conf. Comp. Vis.}, 2017.

\bibitem{YangWXYK18}
Linjie Yang, Yanran Wang, Xuehan Xiong, Jianchao Yang, and Aggelos~K.
  Katsaggelos.
\newblock Efficient video object segmentation via network modulation.
\newblock In {\em Proc. IEEE Conf. Comp. Vis. Patt. Recogn.}, 2018.

\bibitem{adaptis}
Konstantin Sofiiuk, Olga Barinova, and Anton Konushin.
\newblock {AdaptIS}: Adaptive instance selection network.
\newblock In {\em Proc. IEEE Int. Conf. Comp. Vis.}, 2019.

\bibitem{CondInst}
Zhi Tian, Chunhua Shen, and Hao Chen.
\newblock Conditional convolutions for instance segmentation.
\newblock In {\em Proc. Eur. Conf. Comp. Vis.}, 2020.

\bibitem{bodla2017soft}
Navaneeth Bodla, Bharat Singh, Rama Chellappa, and Larry Davis.
\newblock Soft-{NMS}: improving object detection with one line of code.
\newblock In {\em Proc. IEEE Int. Conf. Comp. Vis.}, 2017.

\bibitem{liu2019adaptive}
Songtao Liu, Di~Huang, and Yunhong Wang.
\newblock Adaptive {NMS}: Refining pedestrian detection in a crowd.
\newblock In {\em Proc. IEEE Conf. Comp. Vis. Patt. Recogn.}, 2019.

\bibitem{he2019bounding}
Yihui He, Chenchen Zhu, Jianren Wang, Marios Savvides, and Xiangyu Zhang.
\newblock Bounding box regression with uncertainty for accurate object
  detection.
\newblock In {\em Proc. IEEE Conf. Comp. Vis. Patt. Recogn.}, 2019.

\bibitem{CaiZWLFAC19}
Lile Cai, Bin Zhao, Zhe Wang, Jie Lin, Chuan~Sheng Foo, Mohamed M.~Sabry Aly,
  and Vijay Chandrasekhar.
\newblock Maxpoolnms: Getting rid of {NMS} bottlenecks in two-stage object
  detectors.
\newblock In {\em Proc. IEEE Conf. Comp. Vis. Patt. Recogn.}, 2019.

\bibitem{coordconv}
Rosanne Liu, Joel Lehman, Piero Molino, Felipe~Petroski Such, Eric Frank, Alex
  Sergeev, and Jason Yosinski.
\newblock An intriguing failing of convolutional neural networks and the
  coordconv solution.
\newblock In {\em Proc. Advances in Neural Inf. Process. Syst.}, 2018.

\bibitem{kirillov2019panoptic}
Alexander Kirillov, Ross Girshick, Kaiming He, and Piotr Doll{\'a}r.
\newblock Panoptic feature pyramid networks.
\newblock In {\em Proc. IEEE Conf. Comp. Vis. Patt. Recogn.}, 2019.

\bibitem{wu2018group}
Yuxin Wu and Kaiming He.
\newblock Group normalization.
\newblock In {\em Proc. Eur. Conf. Comp. Vis.}, 2018.

\bibitem{focalloss}
Tsung-Yi Lin, Priya Goyal, Ross Girshick, Kaiming He, and Piotr Doll{\'a}r.
\newblock Focal loss for dense object detection.
\newblock In {\em Proc. IEEE Int. Conf. Comp. Vis.}, 2017.

\bibitem{masklab}
Liang-Chieh Chen, Alexander Hermans, George Papandreou, Florian Schroff, Peng
  Wang, and Hartwig Adam.
\newblock Masklab: Instance segmentation by refining object detection with
  semantic and direction features.
\newblock In {\em Proc. IEEE Conf. Comp. Vis. Patt. Recogn.}, 2018.

\bibitem{wang2020centermask}
Yuqing Wang, Zhaoliang Xu, Hao Shen, Baoshan Cheng, and Lirong Yang.
\newblock Centermask: single shot instance segmentation with point
  representation.
\newblock In {\em Proc. IEEE Conf. Comp. Vis. Patt. Recogn.}, 2020.

\bibitem{coco}
Tsung{-}Yi Lin, Michael Maire, Serge~J. Belongie, James Hays, Pietro Perona,
  Deva Ramanan, Piotr Doll{\'{a}}r, and C.~Lawrence Zitnick.
\newblock Microsoft {COCO:} common objects in context.
\newblock In {\em Proc. Eur. Conf. Comp. Vis.}, 2014.

\bibitem{lvis2019}
Agrim Gupta, Piotr Dollar, and Ross Girshick.
\newblock {LVIS}: A dataset for large vocabulary instance segmentation.
\newblock In {\em Proc. IEEE Conf. Comp. Vis. Patt. Recogn.}, 2019.

\bibitem{Islam2020How}
Md~Amirul Islam, Sen Jia, and Neil D.~B. Bruce.
\newblock How much position information do convolutional neural networks
  encode?
\newblock In {\em Proc. Int. Conf. Learn. Representations}, 2020.

\bibitem{li2019attention}
Yanwei Li, Xinze Chen, Zheng Zhu, Lingxi Xie, Guan Huang, Dalong Du, and
  Xingang Wang.
\newblock Attention-guided unified network for panoptic segmentation.
\newblock In {\em Proc. IEEE Conf. Comp. Vis. Patt. Recogn.}, 2019.

\bibitem{xiong2019upsnet}
Yuwen Xiong, Renjie Liao, Hengshuang Zhao, Rui Hu, Min Bai, Ersin Yumer, and
  Raquel Urtasun.
\newblock {UPSNet}: A unified panoptic segmentation network.
\newblock In {\em Proc. IEEE Conf. Comp. Vis. Patt. Recogn.}, 2019.

\bibitem{cheng2019panoptic}
Bowen Cheng, Maxwell Collins, Yukun Zhu, Ting Liu, Thomas Huang, Hartwig Adam,
  and Liang-Chieh Chen.
\newblock Panoptic-deeplab: A simple, strong, and fast baseline for bottom-up
  panoptic segmentation.
\newblock In {\em Proc. IEEE Conf. Comp. Vis. Patt. Recogn.}, 2020.

\bibitem{yolov3}
Joseph Redmon and Ali Farhadi.
\newblock Yolov3: An incremental improvement.
\newblock {\em arXiv:1804.02767}, 2018.

\bibitem{ssd}
Wei Liu, Dragomir Anguelov, Dumitru Erhan, Christian Szegedy, Scott Reed,
  Cheng-Yang Fu, and Alexander~C Berg.
\newblock Ssd: Single shot multibox detector.
\newblock In {\em Proc. Eur. Conf. Comp. Vis.}, 2016.

\bibitem{refinedet}
Shifeng Zhang, Longyin Wen, Xiao Bian, Zhen Lei, and Stan~Z Li.
\newblock Single-shot refinement neural network for object detection.
\newblock In {\em Proc. IEEE Conf. Comp. Vis. Patt. Recogn.}, 2018.

\bibitem{fpn}
Tsung{-}Yi Lin, Piotr Doll{\'{a}}r, Ross~B. Girshick, Kaiming He, Bharath
  Hariharan, and Serge~J. Belongie.
\newblock Feature pyramid networks for object detection.
\newblock In {\em Proc. IEEE Conf. Comp. Vis. Patt. Recogn.}, 2017.

\bibitem{foveabox}
Tao Kong, Fuchun Sun, Huaping Liu, Yuning Jiang, and Jianbo Shi.
\newblock Foveabox: Beyond anchor-based object detector.
\newblock {\em arXiv:1904.03797}, 2019.

\bibitem{reppoints}
Ze~Yang, Shaohui Liu, Han Hu, Liwei Wang, and Stephen Lin.
\newblock Reppoints: Point set representation for object detection.
\newblock In {\em Proc. IEEE Int. Conf. Comp. Vis.}, 2019.

\bibitem{centernet}
Xingyi Zhou, Dequan Wang, and Philipp Kr{\"a}henb{\"u}hl.
\newblock Objects as points.
\newblock {\em arXiv:1904.07850}, 2019.

\bibitem{mmdetection}
Kai Chen, Jiaqi Wang, Jiangmiao Pang, Yuhang Cao, Yu~Xiong, Xiaoxiao Li,
  Shuyang Sun, Wansen Feng, Ziwei Liu, Jiarui Xu, Zheng Zhang, Dazhi Cheng,
  Chenchen Zhu, Tianheng Cheng, Qijie Zhao, Buyu Li, Xin Lu, Rui Zhu, Yue Wu,
  Jifeng Dai, Jingdong Wang, Jianping Shi, Wanli Ouyang, Chen~Change Loy, and
  Dahua Lin.
\newblock {MMDetection}: Open mmlab detection toolbox and benchmark.
\newblock {\em arXiv:1906.07155}, 2019.

\end{thebibliography}
}

\end{document}